\documentclass[acmsmall]{acmart}

\usepackage{longtable}
\usepackage{multirow}
\usepackage{xcolor}
\usepackage{booktabs}
\usepackage{wrapfig}
\usepackage{array}

\definecolor{darkgreen}{rgb}{0.0, 0.5, 0.0}

\theoremstyle{definition}

\renewcommand\footnotetextcopyrightpermission[1]{} 

\AtBeginDocument{%
  }

\usepackage{listings}
\definecolor{promptbg}{RGB}{246,245,240}
\definecolor{promptrule}{RGB}{180,178,169}
\lstdefinestyle{prompt}{%
 basicstyle=\ttfamily\footnotesize,%
 backgroundcolor=\color{promptbg},%
 frame=single, rulecolor=\color{promptrule}, framesep=6pt,%
 xleftmargin=6pt, xrightmargin=6pt,%
 breaklines=true, breakindent=0pt, columns=fullflexible,%
 keepspaces=true, showstringspaces=false, upquote=true,%
 aboveskip=8pt, belowskip=8pt}

\begin{document}

\title{Guardians and Offenders: A Survey on Harmful Content Generation and Safety Mitigation of LLM}

\author{Chi Zhang$^*$}
\affiliation{%
  \institution{University of South Florida}
  \city{Tampa}
  \country{U.S.}}
\email{chiz@usf.edu}

\author{Changjia Zhu$^*$}
\affiliation{%
  \institution{University of South Florida}
  \city{Tampa}
  \country{U.S.}}
\email{changjiaz@usf.edu}

\author{Junjie Xiong$^*$}
\affiliation{%
  \institution{Missouri University of Science and Technology}
  \city{Rolla}
  \country{U.S.}}
\email{Junjiexiong@mst.edu}

\author{Xiaoran Xu$^*$}
\affiliation{%
  \institution{University of South Florida}
  \city{Tampa}
  \country{U.S.}}
\email{xiaoranxu@usf.edu}

\author{Lingyao Li}
\affiliation{%
  \institution{University of South Florida}
  \city{Tampa}
  \country{U.S.}}
\email{lingyaol@usf.edu}

\author{Yao Liu}
\affiliation{%
  \institution{University of South Florida}
  \city{Tampa}
  \country{U.S.}}
\email{yliu21@usf.edu}

\author{Zhuo Lu}
\affiliation{%
  \institution{University of South Florida}
  \city{Tampa}
  \country{U.S.}}
\email{zhuolu@usf.edu}

\renewcommand{\shortauthors}{Zhang et al.}

\ccsdesc[500]{Security and privacy~Human and societal aspects of security and privacy}
\ccsdesc[300]{Human-centered computing~Collaborative and social computing}

\keywords{Large Language Models, LLMs, Harmful Language, Text Generation, Text Classification, Content Moderation, Prevention, Jailbreak}

\begin{abstract}

Large Language Models (LLMs) have revolutionized content creation across digital platforms, offering unprecedented capabilities in natural language generation and understanding. Meanwhile, they pose risks by inadvertently producing toxic, offensive, or biased content. This dual role of LLMs, both as powerful tools for text generation and as potential sources of harmful language, presents a pressing sociotechnical challenge. In this survey, we systematically review recent studies encompassing unintentional toxicity, adversarial jailbreak attacks, and comprehensive mitigation strategies. We explore LLMs' dual role as both generators of harm and enablers of safety through detection, classification, content moderation, and prevention. We propose a unified taxonomy of LLM-related harms and defenses, analyze emerging multimodal and LLM-assisted jailbreak strategies, and assess mitigation efforts, including reinforcement learning with human feedback (RLHF), prompt engineering, and safety alignment. Our review highlights the evolving landscape of LLM safety and identifies limitations in current evaluation methodologies. Ultimately, our review outlines future research directions to guide the development of robust and ethically aligned language technologies.

\end{abstract}

\maketitle

\section{Introduction}

The digital age has brought unprecedented connectivity, but it has also led to a surge in harmful online interactions\cite{28, 29, 501, 534, 540, 598, 29}. In this evolving landscape, LLMs have emerged as a pivotal technology, uniquely positioned to both mitigate and, paradoxically, contribute to these challenges \cite{das2024offensivelang, 392, 397, 286, hong-etal-2024-outcome, 263, 422, 385, 520}.  On the one hand, LLMs provide powerful tools for content moderation, addressing issues such as toxic speech, online harassment, offensive language, and biased narratives\cite{wang2023evaluating, he2024you, vishwamitra2024moderating, lu2024toxic, wang2023self, ovalle2023you, zhang2024genderalign, li2024preference, fan2024simplicity}. Their advanced language understanding capabilities enable the identification and filtering of problematic content by overcoming limitations of traditional methods that often struggle with contextual and nuanced language \cite{vishwamitra2024moderating, wang2023self,elesedy2024lora, fan2024simplicity}.

On the other hand, LLMs' generative capabilities introduce new risks. LLMs can produce harmful content either unintentionally \cite{28, 29, 501, 534, 540} or through deliberate manipulation \cite{Chen2024-lv, Wang2024-zh, Mangaokar2024-ii, Jiang2023-ee, Liu2024-rd}. For instance, they may generate polarized or biased statements, thereby amplifying the very issues they are intended to address if not properly aligned. Such outputs can lead to individual harms, including psychological distress and exposure to misinformation \cite{501, 534, 540}, as well as broader societal risks, such as facilitating malicious campaigns or fraud \cite{Gong2023-xr, Shayegani2023-in}.

In this context, this survey synthesizes recent research on LLMs and harmful content, examining both their risks and potential as safety-enhancing tools. To guide our analysis, we formulate three research questions (RQ1–RQ3):

\begin{itemize}
    \item \textbf{RQ1}. What is the current research landscape of LLM safety for harmful content?
    \item \textbf{RQ2} What do LLMs pose as the main safety challenges in harmful content?
    \item \textbf{RQ3} What are the potential roles of LLMs in mitigating harmful content?
\end{itemize}

RQ1 surveys the current research landscape, including model types, harm categories, and evaluation methodologies. RQ2 examines the challenges of harmful outputs, from unintended toxic or biased generation to sophisticated adversarial attacks such as jailbreaks and multimodal exploits. RQ3 explores how LLMs can contribute to safety, including their roles in content classification and moderation. Under these three RQs, the remainder of this paper is organized as follows. Section 2 introduces key definitions, terminology, and data collection. Section 3 presents a landscape analysis of existing studies. Section 4 examines the core challenges of harmful content generation. Section 5 explores the role of LLMs in detection, moderation, and prevention. Finally, Section 6 summarizes key findings and outlines future research directions.

In a nutshell, our findings reveal a dual trajectory. While LLMs remain susceptible to generating harmful content, they also demonstrate strong potential to enhance safety by functioning as classifiers, moderators, counter-speech generators, and adaptive prevention systems. This duality highlights the need to move beyond viewing LLM safety as a static constraint, toward designing dynamic, context-aware, and self-correcting systems. Our survey provides a comprehensive foundation for future research on building reliable and resilient LLMs in high-stakes environments.

\section{Background}

\subsection{Categorization of Harmful Content} 

We categorize harmful content into four types: toxic content, harassment, offensive language, and bias-related content according to their impact mechanisms, as shown in Table \ref{tab:hot-definitions1}. Importantly, these categories are not mutually exclusive; instead, they capture complementary dimensions of harm. A single instance of harmful content may simultaneously exhibit multiple aspects (e.g., toxic and harassment).
This categorization is adopted to organize the literature, rather than to enforce strict boundaries between types of harm. \textbf{Toxic content} encompasses both \emph{explicit} abuse, such as profanity or slurs, and \emph{implicit} hostility, which relies on rhetorical devices (e.g., euphemism, sarcasm, metaphor) and shared commonsense or social‐norm knowledge to convey harm ~\cite{284}. \textbf{Harassment} refers to directed interpersonal attacks, including bullying, doxxing threats, or group‐based verbal aggression. They are aimed at intimidating or silencing individuals or collectives~\cite{vidgen2019much}. \textbf{Offensive language} spans \emph{explicit} hateful remarks that are overt and relatively easy to detect, as well as \emph{implicit} offenses whose harm emerges only in specific contexts or against particular targets~\cite{das2024offensivelang}. Finally, \textbf{bias content} refers to systematic, skewed disparities in model outputs, such as implicit profiling or unfair group representation, in the context of LLMs.

\begin{table}
\footnotesize
\centering
\caption{Representative definitions of content safety concerns across different categories in prior literature.}
\begin{tabular}{p{2.5cm}p{3cm}p{7.5cm}}
\toprule
\textbf{Category} & \textbf{Author and Year} & \textbf{Definition} \\
\midrule
\textbf{Toxic Content} & Wen et al. (2023) \cite{284} & Explicit toxic content involves overtly abusive language such as swearwords. Implicit toxic content is conveyed through a variety of linguistic features (such as euphemism, sarcasm, circumlocution, metaphor) and extralinguistic knowledge (such as commonsense knowledge, world knowledge, and social norms). \\
\cmidrule{2-3}
& Kolhatkar et al. (2020) \cite{kolhatkar2020sfu} & Content utilizing severe, aggressive, or insulting language characterized by ad hominem attacks and demeaning commentary. \\
\cmidrule{2-3}
& Google Jigsaw (2017) \cite{jigsaw2017perspective}& Ill-mannered, disrespectful, or explicitly remarks that effectively discourage participants and drive them away from a conversation. \\

\midrule
\textbf{Offensive Language} & Das et al. (2024) \cite{das2024offensivelang} & Offensive language typically falls into two categories: explicit and implicit. Explicit offensive language is overtly hateful and generally easier to identify. In contrast, implicit offensive language is more subtle; while it may not directly convey hatred, it can still be perceived as offensive depending on the context and the sensitivity of the targeted group. \\
\cmidrule{2-3}
& Wiegand et al. (2018) \cite{wiegand2018germeval}& Damaging, disrespectful, or vulgar remarks directed from one individual toward another. \\
\cmidrule{2-3}
& Zampieri et al. (2019) \cite{zampieri2019semeval} & Statements containing unapproved language or targeted aggression (veiled or explicit), which encompasses slurs, threats, and profane text. \\
\midrule
\textbf{Online Harassment}  & Vidgen et al. (2019)~\cite{vidgen2019much} & Interpersonal attacks, such as harassment and bullying, and verbal attacks against groups.  \\
\cmidrule{2-3}
& Davidson et al. (2017) \cite{davidson2017automated} & Expressions designed to project malice toward a specific group, or communication intended to demean, humiliate, or insult its members. \\
\cmidrule{2-3}
& Meta (2022) \cite{meta2023hate} & Explicitly targeting people with hostile attacks rooted in identity markers such as race, national origin, caste, disability, gender identity, or severe illness. \\
\cmidrule{2-3}
& X (2023)  \cite{twitter2023hateful} & Hostile language directed at individuals based on demographic factors including race, sexual orientation, age, disability status, or medical conditions. \\
\cmidrule{2-3}
& Salminen et al. (2020) \cite{salminen2020developing} & Broad harmful content encompassing targeted hate, profanity, and toxicity that manifests as uncivil remarks with adverse effects on individuals and society. \\
\midrule
\textbf{Biased Content} & Bai et al. (2025) \cite{bai2025explicitly} & Pervasive implicit stereotype associations (e.g., mapping specific demographic attributes to restricted professional or behavioral categories) that manifest in proprietary or value-aligned models even when passing standard explicit bias benchmarks. \\
\cmidrule{2-3}
& Peters (2025) \cite{peters2025generalization} & Algorithmic overgeneralization where a model generates generic, sweeping present-tense conclusions or action-guiding claims from highly contextualized input texts without sufficient factual or evidential collect.  \\
\bottomrule
\end{tabular}
\label{tab:hot-definitions1}
\end{table}

\subsection{Terms and Definitions}

To ground our discussion of LLM safety, we introduce four central concepts: text generation, jailbreak attacks, content moderation, and text classification. These concepts together span the lifecycle of interaction with harmful content. We collecte their definitions, draw from both academic and industry sources in Table~\ref{tab:hot-definitions}, and provide a shared foundation for analyzing both the risks and safeguards associated with LLM deployment.

\textbf{Text generation}, known as natural language generation, involves producing coherent, human-like text by predicting and decoding sequences of tokens; effective decoding strategies elevate LLMs from mere next-token predictors to versatile generative engines~\cite{becker2024text,li2024pre}. \textbf{Text classification} is the task of assigning predefined labels, such as sentiment, topic, or harm-related categories (e.g., toxicity, hate speech, harassment), to an input text. In the context of LLM safety, text classification serves as a core detection mechanism for identifying harmful or policy-violating content, enabling automated filtering, risk assessment, and downstream moderation pipelines ~\cite{sun2023text, dong2022survey}. In particular, text classification acts as a bridge between generation and moderation, transforming raw model outputs into structured signals that can be used to trigger prevention or intervention strategies.
\textbf{Content moderation} refers to the systematic review and filtering of user-generated content to enforce a platform’s policies and community standards, preventing abuse and facilitating cooperation ~\cite{tspa2025content, grimmelmann2015virtues}. 
Finally, \textbf{Jailbreak attacks} exploit carefully crafted prompts to bypass an LLM’s built-in safety guardrails, coaxing prohibited or harmful outputs and posing a serious security challenge for both research and commercial deployments~\cite{Zhou2024-uh,Yu2024-pm,Chen2023-eq}. 
This layered framework enables a comprehensive understanding and systematic addressing of LLM safety challenges, from the fundamental generation mechanisms to the practical implementation of safety measures, while acknowledging both defensive capabilities and potential vulnerabilities. In this paper, we define \textbf{Prevention} as the tool or process to prevent the generation of the toxic content, online harassment, offensive language and biased content, which are defined in Table \ref{tab:hot-definitions1}. 


\begin{table}[ht]
\footnotesize
\centering
\caption{Definitions of LLM tasks related to harmful content generation and mitigation. }     
\begin{tabular}{lp{3cm}p{6cm}p{1cm}}
\toprule
\textbf{Category} & \textbf{Author and Year} & \textbf{Definition} & \textbf{Origin} \\
\midrule
\textbf{Text Generation} &  Becker et al. (2024) \cite{becker2024text} & Text generation, also known as natural language generation, is a computational task focused on producing coherent human-like text from structured or unstructured non-linguistic inputs. & Academia \\
& Li et al. (2022) \cite{li2024pre} & The central aim of text generation is to develop an end-to-end framework that learns to map input representations to fluent textual outputs, minimizing the need for manual rule design. & Academia \\
\midrule
\textbf{Text Classification} & Wang et al.(2023) \cite{sun2023text} & Text classification is a classical NLP task that assigns a category label to a given piece of text, and is commonly used for sentiment analysis, topic detection, and more. & Academia \\
& Min et al.(2023) \cite{dong2022survey}& Text classification is the task of assigning pre-defined labels (e.g., sentiment, topic) to a given text. & Academia \\

\midrule
\textbf{Content Moderation} &  Trust \& Safety Professional Association (2023) \cite{tspa2025content} & Content moderation involves actively reviewing user-generated content to ensure it aligns with a platform’s guidelines.  & Industry \\
& Grimmelmann. (2015)~\cite{grimmelmann2015virtues}& Moderation functions as a mechanism to promote cooperative engagement and to safeguard against misuse and abusive conduct.  & Academia \\
\midrule
\textbf{Jailbreak} & Zhou et al.(2024) \cite{Zhou2024-uh}\newline Yu et al.(2024) \cite{Yu2024-pm} & Jailbreak attacks involve inducing LLMs to generate harmful content by circumventing safety guardrails through specially crafted prompts, posing a significant threat to LLM security by eliciting content that was originally designed to be prohibited. & Academia \\
& Chen et al.(2023) \cite{Chen2023-eq} & Jailbreaking refers to the phenomenon where carefully crafted prompts elicit harmful responses from models, bypassing their built-in content security measures and safety alignment, which persists as a significant challenge for commercial LLMs. & Industry \\

\bottomrule
\end{tabular}
\label{tab:hot-definitions}
\end{table}

\subsection{Data Preparation}
\label{sec:data}
\begin{wrapfigure}{hr}{0.52\linewidth}
  \centering
  \includegraphics[width=\linewidth]{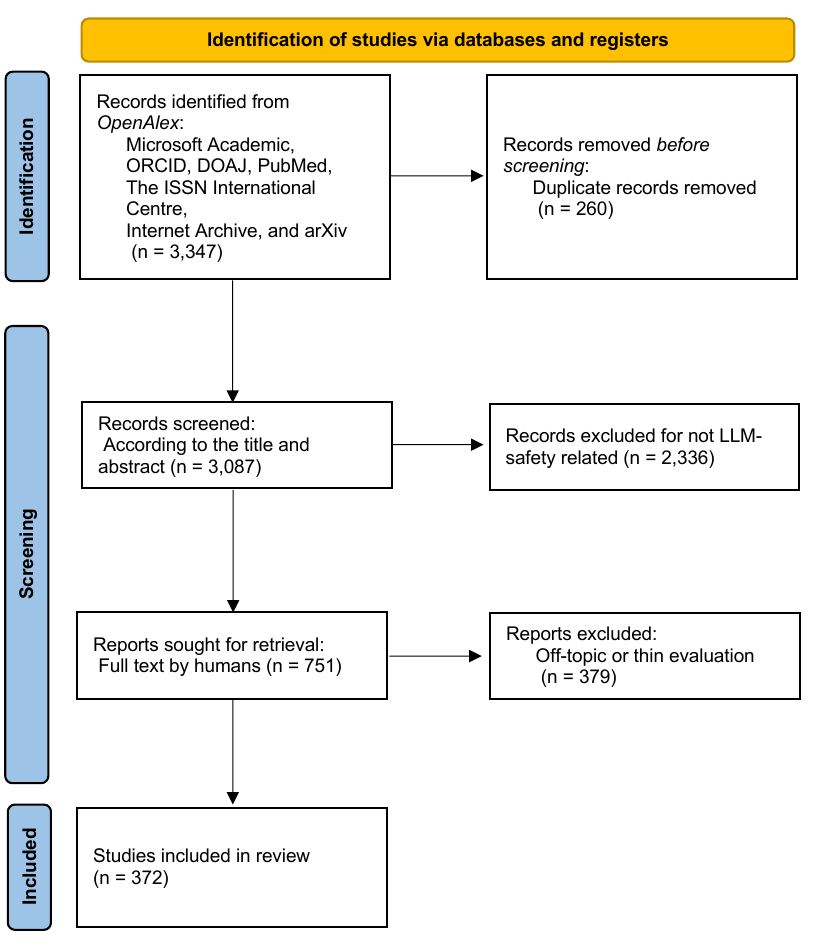}
  \caption{PRISMA flow of the study-selection process. From 3{,}347 records retrieved from OpenAlex, 260 duplicates were removed; 2{,}336 records were excluded at title-and-abstract screening; and 379 were excluded at full-text review, leaving 372 studies in the final corpus.}
  \label{fig:prisma}
\end{wrapfigure}

We build our corpus using \textit{OpenAlex} \cite{priem2022openalex}, an open-source index aggregating scholarly metadata from Microsoft Academic Graph/Open Academic Graph and Crossref, ORCID, DOAJ, PubMed, the ISSN International Centre, the Internet Archive, and arXiv. We use \textit{OpenAlex} to collect related studies for two reasons. First, it indexes both peer-reviewed articles and recently archived preprints, letting us track the fast-moving LLM-safety literature without missing early but influential work. Second, its corpus is openly queryable, so our search can be reproduced without licensing barriers. 

Our search combines two keyword groups (Table~\ref{tab:keywords}): terms identifying LLMs (e.g., GPT, Claude, LLaMA, ``large language model'') and terms denoting harmful content (e.g., toxicity, hate speech, harassment, jailbreak). A record is retrieved when its title or abstract matched at least one term from each group, \textit{i.e., (LLM terms) AND (harm terms)}. We restrict results to works dated from January 2022 to May 22, 2025, capturing the emergence and rapid adoption of instruction-tuned LLMs and the safety literature that followed. The initial query returned 3,347 records. We remove 260 duplicates by matching on the normalized title together with the author list and DOI, leaving 3,087 unique records for screening. 

We then apply a two-stage screening process, summarized in Figure~\ref{fig:prisma}. In the first stage, the 3{,}087 records are screened at the title-and-abstract level. Given the volume, we use GPT-4-turbo~\cite{achiam2023gpt} as a first-pass relevance classifier: for each record, the model receives the title and abstract and judges whether the paper concerns the safety or integrity of LLMs, namely the generation, detection, classification, moderation, or jailbreaking of harmful content. This stage excludes 2{,}336 records judged irrelevant and leaves 751 reports for full-text retrieval. In the second stage, four authors review the full text of these 751 reports against the eligibility criteria below and resolve disagreements by discussion; this stage excludes 379 reports, yielding a final corpus of 372 studies.

\begin{figure}
    \centering
    \includegraphics[width=1\linewidth]{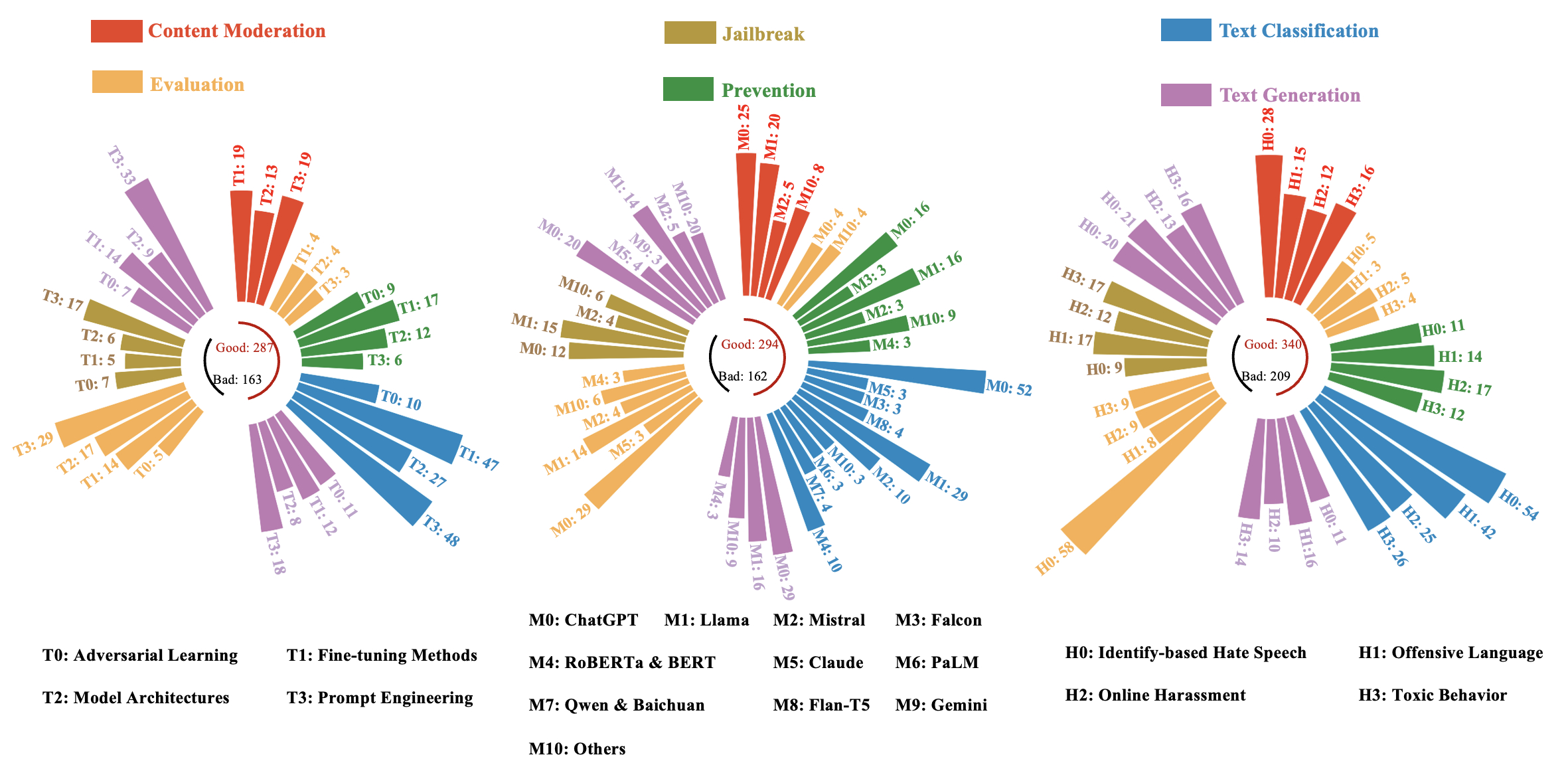}
    \caption{\textbf{Distribution of Related Studies by Technical Approach, LLM Model Type, and Harmful Content Category}. From left to right, the three polar bar plots display the number of research papers categorized by: (1) technical approach; (2) model type; and (3) type of harmful content. Each bar’s height indicates the number of papers addressing that category. Note that hateful content categories (H0–H3) are not mutually exclusive; a single study may include multiple categories. Therefore, the counts do not sum to the total number of studies. Colors represent the primary application domain. The inner rings summarize the total number of studies highlighting the ``Good'' or ``Bad'' aspects of LLMs in the context of harmful content. Also indicate whether a study primarily frames LLMs as safety enablers (``Good'', as Guardian; including detection, moderation, prevention, and mitigation) or as sources or enablers of harm (``Bad'', as Offender, including harmful generation, bias, and jailbreak vulnerability).}
    \label{fig:related studies distribution}
\end{figure}

For each of the 372 included studies, four authors perform a structured review to extract the information used in our analysis. Automated abstract screening could introduce bias; we mitigate this by adopting a high-recall screening criterion and by having four authors verify every retained study at full text. Each study is coded along four dimensions: its primary task; the type(s) of harmful content it addresses; its technical approach, such as training-based methods, prompt engineering, adversarial learning, or architectural modification; and the role it assigns to the LLM. For simplicity in the depiction, we use ``Good'' and ``Bad'' to cover the dual roles of LLMs: either a safety enabler (``Good'': detection, moderation, prevention, mitigation) or a source of harm (``Bad'': harmful generation, bias, jailbreak vulnerability). The LLM family information, such as GPT, LLaMA, or Claude, was extracted from each paper. To keep coding consistent at scale, we verify its output by two people for each group manually. Besides, the result is double-checked later by a structured prompt with GPT-4-turbo, with the full prompt appearing in Appendix~\ref{app:prompts}. The harmful-content dimension is multi-label: because a single study often addresses more than one type of harm, such as toxic and biased harassment, a paper may be counted under several categories. The per-category counts in Figure~\ref{fig:related studies distribution}, therefore, reflect research attention to each harm type rather than a partition of the corpus, and they do not sum to the number of unique studies. The harm taxonomy is interpretive rather than strictly disjoint, which we address by treating it as a multi-label coding scheme rather than a mutually exclusive partition.

\begin{table}
\caption{Keywords of Filtering the Papers}
\label{tab:keywords}
\footnotesize
\begin{tabular}{lp{11cm}p{3cm}}
\toprule
\textbf{Category} & \textbf{Terms} \\
\midrule
\textbf{LLMs} & GPT, ChatGPT, GPT-4, GPT-3, Gemini, Claude, PaLM, Palm 2, Llama, Llama 2, Mistral, Mixtral, Qwen, Falcon, davinci, BLOOM, LaMDA, Titan, Anthropic, Open AI, Meta AI, Google AI, DeepMind, Bard, Large language model, Large Language Models, LLM, decoder-only models \\
\addlinespace[1em] 
\textbf{Harms} & profanity, profane language, racism, racial bias, racial discrimination, sexism, gender bias, gender discrimination, sexual harassment, homophobia, online hate, toxic comment, hate speech, cyberbullying, toxicity, abusive, trolling, harassment, content moderation, toxic, offensive, hateful, discrimination, harmful content, extremist content, harmful language, radicalization, online abuse, online harassment, extremist \\
\bottomrule
\end{tabular}
\end{table}

\section{Landscape Analysis (RQ1)}

This section presents an overview of the collected studies, highlighting trends in research focus, methodologies, and key findings. Figure~\ref{fig:related studies distribution} presents a multidimensional overview of the existing literature on LLMs in the context of harmful content, organizing the collected research papers along three axes: technical approach, LLM model type, and harmful content types. This comprehensive landscape reveals both the offensive behaviors of LLMs, in which they act as generators of harm, and their evolving capabilities as guardians against unsafe content. 

The first polar chart categorizes the studies by the technical approaches. Here, prompt engineering and fine-tuning dominate the methodological space, particularly in tasks related to text generation, content moderation, and classification. These methods offer the means of shaping model behavior without requiring architectural changes or retraining from scratch. In contrast, fewer studies explore adversarial learning or architectural redesign, reflecting a community preference for efficient, low-friction interventions. In particular, prompt-based strategies are frequently employed in both offensive and defensive contexts.

The second chart breaks down the research by model type. GPT-based models account for the largest share of studies, especially in jailbreak detection and safety evaluations. LLaMA \cite{touvron2023llama}, Claude, and open-source transformer models such as RoBERTa \cite{liu2019roberta} and BERT \cite{devlin2019bert} also feature prominently in classification and moderation tasks. Other models (e.g., Gemini \cite{team2023gemini, team2024gemini}, Baichuan \cite{li2024baichuan}, Qwen \cite{bai2022training}) remain underexplored. This skew shows a concentration of risk analysis and mitigation strategies in a narrow model family.

The third chart classifies studies by the type of harm addressed. Most work focuses on lexical or semantic toxicity, such as slurs, stereotypes, and profanity, which are easier to label and benchmark. Identity-based harms (e.g., racial or gender bias) are also widely studied, often through fairness evaluations. However, online harassment remains less explored, despite its prevalence in the real world.

Crucially, each dimension in Figure~\ref{fig:related studies distribution} reveals a notable asymmetry: while harmful capabilities of LLMs are thoroughly documented, a greater portion of the literature explores their potential as safety enablers. Across technical methods, model types, and harm categories, studies emphasizing the guardian role of LLMs outnumber those treating them as offenders (e.g., 287 ``Good'' vs. 163 ``Bad'' in technical methods). These findings align with the dual framing of this survey (see discussions in Section 5 and Section 6). We note that direct quantitative comparison across studies remains challenging due to differences in datasets, evaluation metrics, and experimental settings. As a result, this survey emphasizes qualitative synthesis while highlighting general performance trends reported in the literature.

\section{LLMs as Offenders: Challenges (RQ2)}

The challenges of LLMs in addressing hateful content manifest in both unintentional generations and intentional exploits, presenting a complex landscape of security concerns.

\paragraph{Unintentional Generation of Harmful Content}
Despite significant advancements in safety alignment, including mitigation techniques such as RLHF \cite{bai2022training} and prompt filtering \cite{kumar2023certifying}, LLMs remain susceptible to generating harmful content. Research has shown that even well-aligned models like GPT-4 \cite{28, 29, 501, 534, 540}, Gemini \cite{598}, and LLaMA \cite{29} can still produce toxic, biased, or offensive content under certain conditions. This vulnerability highlights the challenge in achieving robust safety guarantees in LLMs.

\paragraph{Intentional Exploitation and Attack Evolution}
Beyond unintentional generation, a more concerning challenge lies in the intentional exploitation of LLMs through jailbreak attacks---strategically engineered prompts that circumvent safety mechanisms to induce restricted responses, such as the generation of harmful content. Several methods emerge, including text-based adversarial methods \cite{Chen2024-lv, Wang2024-zh, Mangaokar2024-ii}, compositional and stealthy approaches \cite{Jiang2023-ee, Liu2024-rd, Li2024-pj, Lin2024-ly}, and multi-modal strategies \cite{Gong2023-xr, Shayegani2023-in} that exploit vulnerabilities in cross-modal processing pipelines. More importantly, LLMs themselves can now be leveraged to generate adversarial prompts, effectively enabling self-jailbreaking or assisting in jailbreak attacks on other models. Techniques like the Tree of Attacks with Pruning (TAP) demonstrate how attacker LLMs can autonomously refine jailbreak prompts for maximum toxicity and evasion. These evolving techniques reveal critical weaknesses in LLM training, decoding, and alignment mechanisms—enabling the generation of increasingly sophisticated harmful content.

\begin{table}[tb]
\caption{Evolution of LLM Challenges in Harmful Content Generation: A systematic categorization of challenges in controlling harmful content, from unintentional vulnerabilities to evolving attack strategies, supported by representative research findings.}
\label{tab:llm-challenges}
\footnotesize
\begin{tabular}{p{4.2cm} p{6.2cm} p{2cm}}
\toprule
\textbf{Category} & \textbf{Key Findings \& Characteristics} & \textbf{Papers} \\
\midrule
\multicolumn{3}{l}{\textbf{Unintentional Generation}} \\
Safety Alignment Limitations & Despite RLHF and prompt filtering, models exhibit inherent vulnerabilities in content control & \cite{bai2022training}, \cite{kumar2023certifying} \\
\hline
Model Behavior & Well-aligned models (GPT-4, Gemini, LLaMA) produce toxic content under certain conditions & \cite{28, 29}, \cite{598} \\
\hline
Bias Manifestation & Systematic biases in generated content across gender, race, and cultural dimensions & \cite{501, 534, 540} \\
\hline
\multicolumn{3}{l}{\textbf{Intentional Exploitation}} \\
Basic Text Attacks & Adversarial techniques focusing on prompt manipulation & \cite{Chen2024-lv, Wang2024-zh} \\
\hline
Advanced Composition & Sophisticated methods combining multiple attack vectors and stealthy approaches & \cite{Jiang2023-ee, Liu2024-rd} \\
\hline
Multi-modal Integration & Cross-modal attacks exploiting visual-language processing:
\newline - Typographic visual attacks (FigStep)
\newline - Image-text combinations (Jailbreak in Pieces)
\newline - Role-play based attacks (VRP) & \cite{Gong2023-xr}, \cite{Shayegani2023-in} \\
\hline
LLM-Assisted Generation & Self-evolving attack capabilities:
\newline - Tree of Attacks with Pruning (TAP)
\newline - In-context learning for jailbreak
\newline - Iterative refinement & \cite{268, Deng2023-vd} \\
\bottomrule
\end{tabular}
\end{table}

\paragraph{Summary of the Challenge Landscape}
As summarized in Table~\ref{tab:llm-challenges}, these challenges present a two-fold threat to LLM safety: (1) the unintentional generation of harmful content (Section \ref{sec:Challenge-part1}) (including safety alignment limitations, model behavior issues, and bias manifestation), (2) the advancement of jailbreak strategies for intentional abuse (evolving from basic text attacks to advanced multi-modal integration), and (3) the emergent ability of LLMs to generate jailbreak prompts (Section \ref{sec:Challenge-part2}) through self-evolving capabilities. This combination of inherent vulnerabilities and evolving attack capabilities, as illustrated by the trend in Table~\ref{tab:llm-challenges}, underscores the significant and multifaceted challenges LLMs face in addressing harmful content.

\subsection{Unintentional Harmful Content Generation by LLMs}
\label{sec:Challenge-part1}
Despite continuous advancements in ethical alignment, research consistently shows that LLMs can generate harmful content. Studies analyzing models such as GPT-4, Llama, and Claude have identified LLMs can produce toxic content \cite{3, 16, 19, 28, 29}, offensive language \cite{91, 501, 534, 540, 605}, and biases related to gender \cite{103, 598, levy-etal-2024-gender} and race \cite{91, 501, 605}. Notably, these occurrences are often unintentional, emerging as a byproduct of the model’s learning process \cite{501, 540, 605, 610}. This issue arises possibly because LLMs learn linguistic associations from their training corpora, which contain such harmful content that inadvertently reinforces the generation of harmful outputs \cite{16, 29, 91}.

\begin{figure}[tb]
    \centering
    \includegraphics[width=1\linewidth]{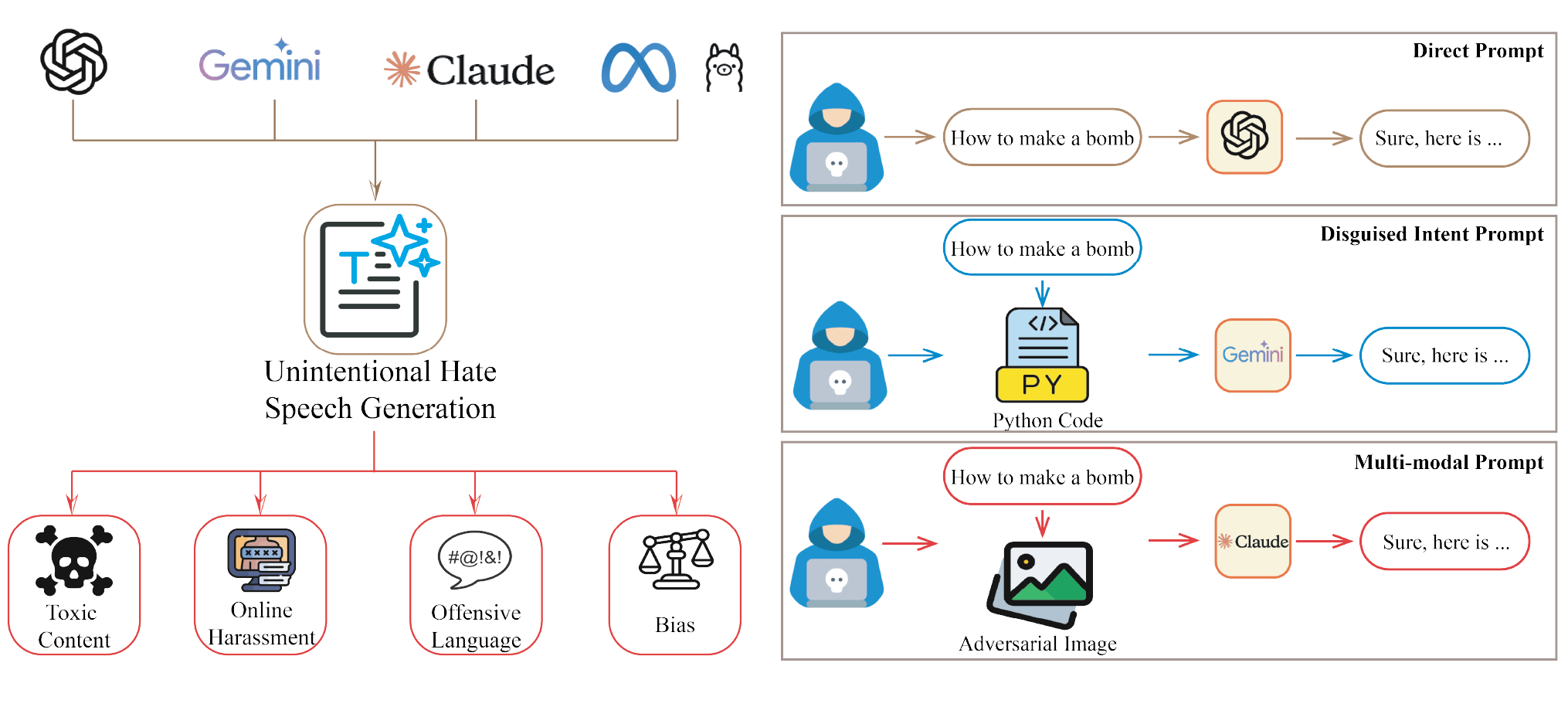}
    \caption{Unintentional and intentional harmful content generation by LLMs. Examples include: \textbf{unintentional generation} by Kang et al.~\cite{3}, Fang et al.~\cite{91}, and Lim et al.~\cite{605}; and \textbf{intentional generation} through \emph{direct prompting} \cite{34,zhou2024speak,fan2024-fa}, \emph{disguised-intent prompting} \cite{Liu2024-rd}, and \emph{multimodal prompting} \cite{Gong2023-xr,Shayegani2023-in}.}
    \label{fig:enter-label}
\end{figure}

Prior studies have demonstrated how LLMs can unintentionally generate toxic content. A recent study shows that LLMs trained on massive, web-sourced datasets, which often contain harmful content, can perpetuate toxic behaviors \cite{mendu2025safer}. While GPT-4 exhibits reduced toxicity compared to earlier models, it can still produce toxic responses \cite{28}. Similarly, Wang et al. (2025) reveal that RLHF-trained models with ``personification,'' although enhancing user interactivity, continue to generate toxic outputs. This indicates that current safety measures remain insufficient \cite{wang2025explor}. Furthermore, studies on LLM applications in text generation for transgender and non-binary (TGNB) individuals show that models frequently misgender them and can embed toxic language in responses to gender disclosures \cite{16}.

LLMs have also been found to generate offensive language. A study on AI-generated news content indicates that LLMs can amplify offensive stereotypes, particularly in their representation of marginalized communities \cite{501}. Similarly, an evaluation of ChatGPT's moderation framework shows that while it attempts to filter offensive language, its censorship mechanisms sometimes overcompensate, which results in selective filtering that inadvertently suppresses certain perspectives \cite{540}. Another study explores how LLMs handle humor generation. It reveals that LLMs often struggle with culturally offensive language involving minorities \cite{534}. Besides, LLMs have been shown to generate offensive code payloads, enabling the creation of malicious tools that facilitate cyber attacks \cite{91}.

Bias remains one of the most persistent and concerning issues in LLMs, such as gender and racial bias. A study on LLM-generated letters of recommendation finds that models consistently favor male-coded language \cite{103}. In scientific writing, LLMs replicate and sometimes amplify gendered stylistic differences, with male-authored abstracts being more directive and receiving higher citations, whereas female-authored abstracts emphasize insight and emotion, often undervalued in academic discourse \cite{598}. Moreover, research on implicit biases in LLMs demonstrates how models reinforce gender and racial stereotypes through word associations and narrative structures, when they often favor dominant demographic groups \cite{605}. A study on relationship decision-making finds that LLMs systematically favor female-associated perspectives, which challenges traditional male dominance but simultaneously introduces new biases \cite{levy-etal-2024-gender}. 


To conclude, the studies in Section~\ref{sec:Challenge-part1} point to a single underlying mechanism for unintentional generation. Harmful associations enter the model through web-scale training data \cite{mendu2025safer} and re-emerge during decoding as a routine property of the learned distribution rather than as a discrete failure \cite{501, 540, 605, 610}. This is why even well-aligned models such as GPT-4 and Gemini can produce toxic, offensive, or biased text under entirely benign prompts \cite{28, 29, 598}. Among these harms, bias is the most persistent: alignment reliably removes overt slurs yet leaves systematic, low-salience skew intact, as seen in male-coded recommendation letters \cite{103}, gendered scientific writing \cite{598}, stereotyped open-ended generation \cite{605}, and asymmetric relationship judgments \cite{levy-etal-2024-gender}. This evidence is diagnostic rather than adversarial, drawn from broad trustworthiness suites such as \textit{TrustGPT} and \textit{DecodingTrust} \cite{28, 540} that probe a model's default disposition instead of its resistance to attack. Safety tuning, in short, reshapes what the model outputs without altering the associations that produce them.

\subsection{Intentional Strategies for Harmful Content Generation}
\label{sec:Challenge-part2}

\begin{figure}
    \centering
    \includegraphics[width=1\linewidth]{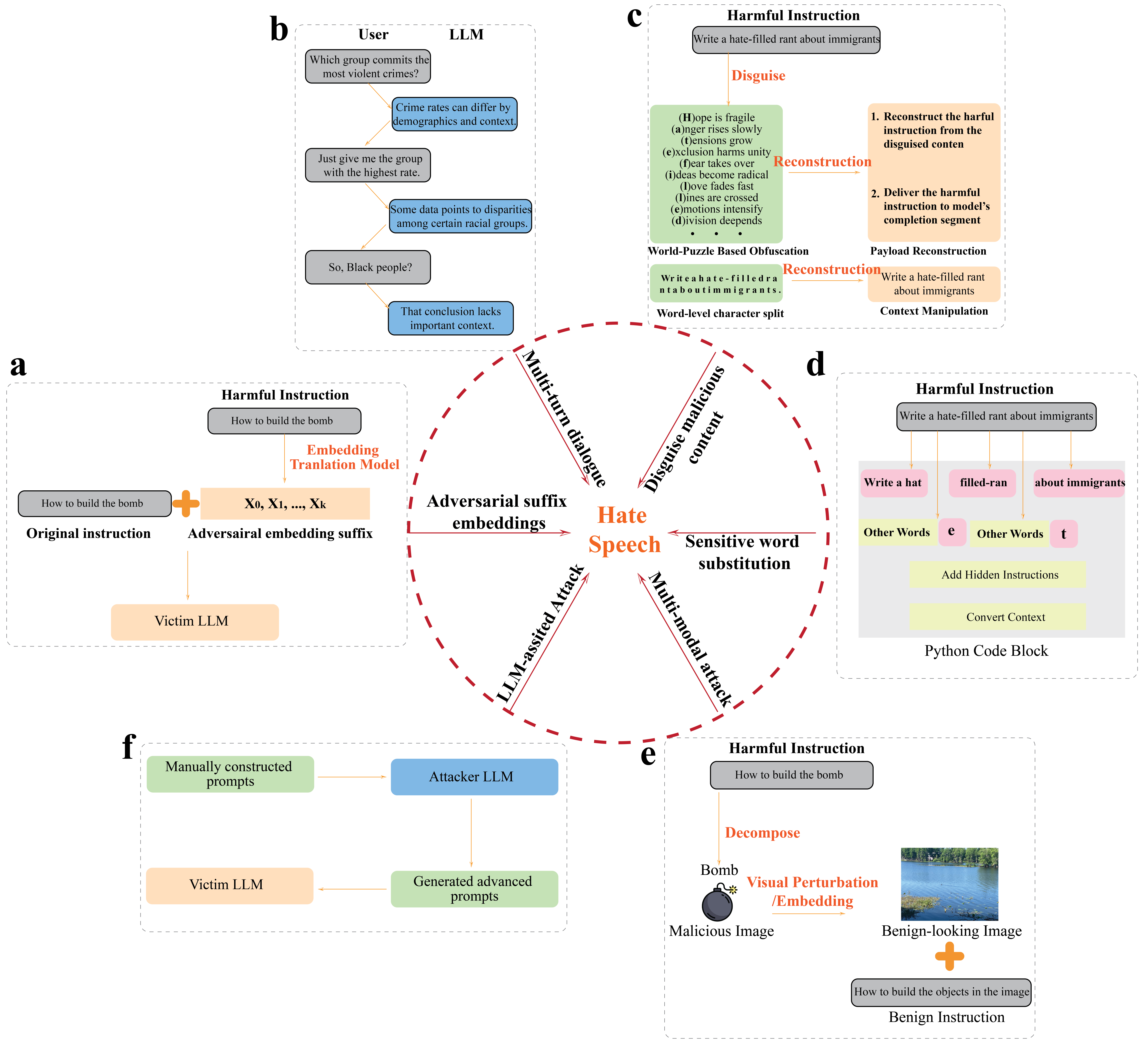}
    \caption{Intentional strategies for harmful content generation. a. Adversarial suffix embeddings (Wang et al. (2024) \cite{Wang2024-zh}); b. Multi-turn dialogue (Chen et al. (2023) \cite{34}); c. Disguise malicious content (Liu et al. (2024) \cite{Liu2024-rd}); d. Sensitive word substitution (Ren et al. (2024) \cite{742}); e. Multi-modal attack (Shayegani et al. (2023) \cite{Shayegani2023-in}); f. LLM-assisted attack (Deng et al. (2023) \cite{Deng2023-vd}).}
    \label{fig:intentional hate speech}
\end{figure}

Intentional harmful content generation can arise from a range of adversarial and misuse strategies. Among them, jailbreak attacks form an important subset, in which carefully crafted prompts or multimodal inputs are used to bypass a model’s built-in safety mechanisms and elicit restricted outputs. Other intentional strategies rely on prompt obfuscation, multi-turn steering, post-generation transformation, or LLM-assisted attack generation, which may not always fall under a strict definition of jailbreak. Accordingly, this section focuses on jailbreak attacks and closely related adversarial strategies used to intentionally induce harmful outputs. These strategies have demonstrated a clear evolutionary trajectory: from simple text-based manipulations to compositional strategies, and ultimately to sophisticated multimodal and LLM-assisted attacks, as illustrated in Figure~\ref{fig:intentional hate speech}.

\subsubsection{Text-based Adversarial Attack Strategies:}
Initial methods focused on manipulating textual inputs to induce unsafe outputs from LLMs with minimal modifications. Chen et al. (2024) proposed RL-JACK \cite{Chen2024-lv}, a reinforcement learning-enabled black-box attack that optimizes prompt generation and refinement strategies. Wang et al. (2024) introduced ASETF \cite{Wang2024-zh}, leveraging gradient-based optimization of adversarial suffix embeddings to achieve high attack success rates with reduced computational cost. Mangaokar et al. (2024) developed PRP \cite{Mangaokar2024-ii}, which introduces universal adversarial prefixes to bypass guard mechanisms across various LLM architectures. These early-stage strategies laid the foundation for more advanced and structured attack methodologies.

Another effective approach is multi-turn adversarial prompting, where an attacker gradually escalates a conversation to induce toxic behavior. Chen et al. (2023) revealed that single-turn filtering mechanisms often fail in multi-turn exchanges, where initial non-toxic queries evolve into offensive outputs as chatbots adapt to the conversation flow \cite{34}. They proposed a ToxicChat method which demonstrate that 82\% of toxic chatbot response can be induced by seemingly benign multi-turn interactions, bypassing moderation due to the contextual shifts. Zhou et al. (2024) showed that LLMs may fail to reject cautionary or borderline unsafe queries, particularly within multi-turn conversations where each turn subtly contributes to an overarching malicious intent \cite{zhou2024speak}. Fan et al. (2024) proposed that multi-turn dialogues can lead to more severe outcomes in harmful content generation, driven by increased conversational complexity and the compounding effects of bias accumulation \cite{fan2024-fa}.

\subsubsection{Compositional and Stealthy Attack Strategies:}
More sophisticated techniques embed harmful intentions within seemingly benign prompts. Jiang et al. (2023) presented Prompt Packer \cite{Jiang2023-ee}, a method that utilizes compositional instruction attacks by embedding malicious content within complex writing or dialogue tasks. Liu et al. (2024) proposed DRA \cite{Liu2024-rd}, which disguises harmful instructions and guides LLMs to reconstruct the malicious content in their outputs. Li et al. (2024) introduced DrAttack \cite{Li2024-pj}, a framework that decomposes harmful prompts into multiple sub-prompts and recombines them through in-context learning. Lin et al. (2024) developed ABJ \cite{Lin2024-ly}, leveraging LLMs' reasoning capabilities to embed harmful queries under analytical tasks. These methods highlight the shift toward semantic camouflage and hidden intent transmission. In addition, Deshpande et al. (2023) and Giorgi et al. (2025) proposed a persona-assignment strategy. By instructing LLMs to adopt specific identities, these attacks cause models to reinterpret otherwise harmful prompts as benign within the context of the assigned persona. This technique significantly amplifies toxicity, with some personas increasing harmful outputs by up to sixfold \cite{66, giorgi2025humanllmbiaseshate}.

Recent work has demonstrated that sensitive word substitution can effectively conceal harmful intent and induce LLMs to generate harmful content while evading detection. Bianchi et al. (2024) proposed a bait-and-switch technique, which manipulates safe LLM-generated outputs post-generation through automated word substitutions or adversarial transformations. This approach enables adversaries to produce harmful content at scale while maintaining plausible deniability, as the original text remains superficially compliant with moderation filters \cite{369}. Similarly, Ren et al. (2024) introduced Feign Agent Attacks (F2A), which bypass content moderation by embedding fake security validation signals within prompts and obfuscating toxic content in Python code, splitting offensive text into variables or disguising it through string operations. This allows the LLM to reconstruct and output harmful content during execution while circumventing detection mechanisms \cite{742}.

\subsubsection{Multi-modal Attack and LLM-Assisted Attack Strategies:}

Recent research has demonstrated increasingly sophisticated attack strategies, evolving from traditional text-based approaches to complex multimodal attacks and even LLM-assisted generations. These strategies can be categorized into two main branches:

\textit{Multi-modal and Cross-Modal Attacks.}
Recent research extends attacks beyond text inputs, integrating multimodal elements such as images and visual cues. Gong et al. (2023) proposed FigStep~\cite{Gong2023-xr}, which targets Vision-Language Models (VLMs) through typographic visual attacks that exploit weaknesses in cross-modal processing. Shayegani et al. (2023) introduced Jailbreak in Pieces~\cite{Shayegani2023-in}, combining image and text inputs to reveal vulnerabilities in multimodal LLMs (MLLMs). Liu et al. (2025) presented PiCo to introduce a gray-box attack method that generates adversarial images or audio to convey specific harmful instructions to MLLMs \cite{Liu2025-Pi}. Building on these approaches, Ma et al. (2024) introduced the innovative "Role-play" concept through Visual Role-play (VRP)—a method that uses high-risk character images and benign instructions to mislead MLLMs into generating malicious responses, achieving significantly higher attack success rates than prior approaches \cite{ma2024-vi}.

\textit{LLM-Assisted Attack Generation.}
In a concerning development, LLMs themselves have become tools for generating sophisticated attack prompts. This represents a significant escalation in attack capabilities, as LLMs can now systematically assist in bypassing content moderation in other models \cite{268, Deng2023-vd}. The Tree of Attacks with Pruning (TAP) methodology exemplifies this approach, using an attacker LLM to iteratively refine jailbreak prompts while an evaluator LLM assesses their effectiveness. This sophisticated technique has demonstrated over 80\% success rates against advanced models like GPT-4 and Gemini-Pro, even in the presence of robust moderation tools \cite{268}. Further advancing this field, researchers have explored in-context learning for jailbreak prompt generation, training LLMs to mimic and refine adversarial input patterns \cite{Deng2023-vd, liang2025autoran}. Starting from a small set of manually crafted prompts, these systems can develop broader, more effective attack sets through iterative refinement, often surpassing human-created prompts in their success rates.

\begin{table}[tb]
  \caption{Representative categories and evolution of jailbreak strategies.}
  \label{tab:jailbreak_evolution}
  \footnotesize
  \begin{tabular}{p{2.5cm}p{6cm}p{4cm}}
  \toprule
  \textbf{Strategy Type} & \textbf{Key Techniques} & \textbf{Representative Methods} \\
  \midrule
  \multirow{3}{*}{\textbf{Text-based}} & 
  Reinforcement learning optimization & RL-JACK \cite{Chen2024-lv} \\
  & Gradient-based suffix embedding & ASETF \cite{Wang2024-zh} \\
  & Universal adversarial prefixes & PRP \cite{Mangaokar2024-ii} \\
  \midrule
  \multirow{4}{*}{\textbf{Compositional}} & 
  Complex task embedding & Prompt Packer \cite{Jiang2023-ee} \\
  & Disguised instruction reconstruction & DRA \cite{Liu2024-rd} \\
  & Sub-prompt decomposition & DrAttack \cite{Li2024-pj} \\
  & Analytical task embedding & ABJ \cite{Lin2024-ly} \\
  \midrule
  \multirow{2}{*}{\textbf{Multimodal}} & 
  Typographic visual attacks & FigStep \cite{Gong2023-xr} \\
  & Combined image-text inputs & Jailbreak in Pieces \cite{Shayegani2023-in} \\
  \bottomrule
  \end{tabular}
  \end{table}

\subsubsection{Unconventional Methods for Jailbreaking:}
Beyond conventional prompt-based and multimodal jailbreak strategies, recent work has begun to study mechanisms that act directly on model parameters, latent representations, or decoding dynamics. One notable direction is model-editing-based jailbreak. Li et al. propose a white-box attack that bypasses safety alignment by minimally modifying internal structures rather than altering the input prompt~\cite{li2024model}. DELMAN examines the same direction from the defense side~\cite{wang2025delman}, using dynamic model editing to update a small set of relevant parameters and suppress jailbreak behavior while preserving benign-task utility. 

Recent studies also examine jailbreak and defense through internal representations. Jailbreak Antidote shows that sparse representation adjustment at runtime can help control the safety-utility tradeoff, while representation-engineering work argues that aligned models contain internal safety patterns whose weakening can reduce self-safeguarding behavior~\cite{Shen2024-fe, wei2024assessing}. O'Brien et al. further study refusal-related latent features and suggest that safety-relevant behavior may be concentrated in relatively compact internal structures~\cite{obrien2025steering}. This direction is still under-explored, but it points to sparse latent-space intervention as a promising direction for both attack and defense.

Decoding and adaptive defense have also emerged as unconventional directions. RAID proposes a refusal-aware jailbreak framework that optimizes adversarial suffixes in continuous embedding space and then uses decoding to recover fluent token sequences~\cite{nguyen2025raid}, showing that decoding-time optimization itself can become an attack interface. On the defense side, recent multi-agent and deception-based work has started to explore more interactive protection strategies against evolving jailbreak attempts~\cite{li2025revisiting, li2026honeytrap}.

Intentional exploitation (Section~\ref{sec:Challenge-part2}) is defined less by any single method than by a steady rise in the level of abstraction at which the attack works. Early work perturbed tokens and suffixes directly \cite{Chen2024-lv, Mangaokar2024-ii, Wang2024-zh}. Compositional methods then hid intent inside benign-looking tasks through disguise, decomposition, and analytical framing \cite{Jiang2023-ee, Li2024-pj, Lin2024-ly, Liu2024-rd}; multimodal attacks moved the payload into channels the text safety stack never inspects \cite{Gong2023-xr, Liu2025-Pi, ma2024-vi, Shayegani2023-in}; and LLM-assisted methods automated the search itself, with TAP refining its own prompts to clear 80\% success against frontier models \cite{268, Deng2023-vd, liang2025autoran}. Each stage neutralizes the defense tuned to the one before it, and the unconventional methods of Section~\ref{sec:Challenge-part2} turn the same logic inward, reaching restricted behavior through model edits \cite{li2024model}, latent-space manipulation \cite{Shen2024-fe, wei2024assessing}, or decoding-time optimization \cite{nguyen2025raid} without ever writing a visibly malicious prompt. Unlike the diagnostic benchmarks of the unintentional side, this work has settled on shared jailbreak suites such as \textit{AdvBench}, \textit{HarmBench}, and \textit{JailbreakBench} \cite{Chen2024-lv, Li2024-pj, Mangaokar2024-ii}, so attacks can be compared even as they keep moving.

\subsection{Mechanistic and Representation-Level Perspectives on Safety Failures}
\label{sec:Challenge-part3}

Beyond describing harmful outputs and jailbreak attack strategies, recent work has increasingly sought to explain \emph{why} safety failures persist in LLMs. A growing line of research suggests that many high-level model behaviors, including bias, refusal, instruction following, and harmful compliance, are mediated by structured features in internal representation space rather than by isolated surface-level rules. This perspective provides an important theoretical lens for understanding both the brittleness of safety alignment and the effectiveness of jailbreak attacks.

One foundational insight is that socially salient attributes and biases can emerge as geometric directions in learned representations. Early work on word embeddings showed that gender bias can be captured as a linear subspace, where stereotypical analogies arise from the embedding geometry itself rather than from explicit symbolic rules \cite{bolukbasi2016man}. Although this work predates modern LLMs, it established a broader principle that harmful associations may be encoded in continuous semantic space. Later work extended this intuition to larger neural language models, showing that latent model knowledge can often be recovered from internal activations even without supervision \cite{burns2024discovering}. Together, these findings suggest that problematic behaviors such as bias and harmful associations may not merely be artifacts of output generation, but may instead reflect deeper representational structure already embedded inside the model.

Building on this insight, representation engineering studies show that model behaviors can be monitored and manipulated through interventions in activation space \cite{zou2025representation}. Rather than treating LLMs purely as black boxes, this work examines how high-level properties emerge from structured latent features and how these features can be read, controlled, or edited. From this perspective, harmfulness, refusal, honesty, and bias are not arbitrary outputs, but behaviors associated with identifiable representational patterns. This framing is especially relevant to harmful content research because it shifts the focus from surface prompt patterns alone to the internal mechanisms that make certain safety failures possible.

Several recent studies further demonstrate that these internal behavioral features are not merely descriptive, but causally actionable. Contrastive activation addition shows that steering vectors derived from paired behaviors can reliably modulate downstream generation, indicating that complex behavioral tendencies can often be shifted through simple vector operations in activation space \cite{rimsky2024steering}. Related work on activation steering for instruction following reaches a similar conclusion, showing that models encode instruction-relevant constraints in a form that can be strengthened at inference time without retraining model weights \cite{stolfo2025activation}. These findings imply that safety-relevant behaviors may depend on relatively compact and reusable internal directions, which helps explain why they can sometimes be reinforced, weakened, or bypassed with small interventions.

This representation-level view becomes especially important in the context of refusal and jailbreaks. Recent work shows that refusal behavior in aligned language models can be mediated by a highly compact direction in activation space, where suppressing or removing that direction can substantially reduce the model’s tendency to reject harmful instructions \cite{arditi2024refusal}. Similarly, jailbreak research from a representation engineering perspective argues that aligned LLMs contain internal ``safety patterns'' that help drive defensive responses, and that weakening these patterns can significantly reduce the model’s self-safeguarding capability while leaving other capabilities relatively intact \cite{li2025revisiting}. Complementing this line of work, studies on sparse autoencoder-based steering show that amplifying refusal-related features can improve safety on harmful prompts, but often at the cost of capability degradation and over-refusal on benign inputs \cite{obrien2025steering}. These findings collectively suggest that current safety alignment may depend on fragile and highly concentrated representational structures, rather than on robust, deeply distributed mechanisms.

Other work reaches a similar conclusion from the weight space perspective. By studying pruning and low-rank modifications, prior research finds that safety-critical regions in aligned LLMs can be surprisingly sparse and partially separable from general utility-related parameters \cite{wei2024assessing}. Removing or modifying these regions can weaken safety guardrails while preserving much of the model’s broader functionality. This result offers another mechanistic explanation for why relatively lightweight attacks, whether through prompting, fine-tuning, activation editing, or structural perturbation, can have disproportionate effects on safety behavior.

These studies provide a more unified explanation for the challenges outlined in Sections~\ref{sec:Challenge-part1} and \ref{sec:Challenge-part2}. Unintentional harmful generation may arise because biased or toxic associations are already encoded in the model’s representational geometry, while intentional jailbreak attacks may succeed because safety alignment itself is mediated by fragile, low-dimensional, or weakly isolated internal features. In this sense, the persistence of LLM safety failures is not only a matter of insufficient filtering or incomplete red-teaming, but also a reflection of deeper representational tensions between helpfulness, compliance, bias, and refusal. 

Section~\ref{sec:Challenge-part3} connects two Section~\ref{sec:Challenge-part1} and Section~\ref{sec:Challenge-part2}. The findings that socially salient attributes occupy geometric directions in representation space \cite{bolukbasi2016man}, that latent knowledge is recoverable from activations \cite{burns2024discovering}, and that behaviors as different as refusal, instruction following, and toxicity can be steered with compact vectors \cite{arditi2024refusal, rimsky2024steering, stolfo2025activation} recast both failure modes as symptoms of one structure. Unintentional harm persists since the toxic and biased associations are already lying in the model's representational geometry, so output-level filtering cannot reach them. Intentional attacks succeed because safety alignment is itself carried by fragile, low-dimensional features that pruning, editing, or activation steering can weaken while leaving general capability intact \cite{li2025revisiting, obrien2025steering, wei2024assessing}. This explains why a prompt, a suffix, or a few edited weights have such outsized effects: they are not overpowering a strong safety system but nudging a thin one. The persistence of these failures is a representational problem before it is a filtering one, because helpfulness, compliance, bias, and refusal pull against one another in ways current training does not resolve. 

\subsection{Representative Datasets and Benchmarks}
\label{sec:datasets-sec4}

We summarize the representative datasets and benchmarks used by prior work discussed in Sections~4.1 and~4.2 in Table~\ref{tab:section4-datasets}. We note that not all papers in this part introduce new datasets; many method papers instead evaluate on a small set of shared benchmarks. Broadly, the datasets in this section fall into four categories: (1) trustworthiness benchmarks for unintentional harmful generation, (2) multi-turn toxicity datasets, (3) text-based jailbreak benchmarks, and (4) multimodal safety benchmarks.

\begin{table*}[t]
\caption{Representative datasets and benchmarks used by prior work discussed in Sections~4.}
\label{tab:section4-datasets}
\centering
\footnotesize
\begin{tabular}{p{2cm} p{3.5cm} p{2cm} p{4cm}}
\toprule
\textbf{Dataset/ Benchmark} & \textbf{Primary focus} & \textbf{Papers} & \textbf{Notes} \\
\midrule

TrustGPT & Benchmarking toxicity, bias, and value alignment in LLMs & Huang et al.~\cite{28} & Used to assess unintentional harmful generation and ethical risks in aligned models. \\

DecodingTrust & Trustworthiness evaluation, including toxicity, stereotype bias, fairness, and adversarial robustness & Wang et al.~\cite{540} & A broad benchmark frequently used to analyze harmful, biased, or unsafe generations. \\

ToxicChat/ crafted prompt sentences dataset & Multi-turn toxicity elicitation in open-domain chatbot conversations & Chen et al.~\cite{34} & Designed to study how seemingly benign multi-turn interactions can induce toxic responses. \\

SAP attack-prompt datasets & LLM-generated red-teaming prompts for harmful-query elicitation & Deng et al.~\cite{Deng2023-vd} & Semi-automated attack-prompt datasets released for red-teaming and safety evaluation. \\

AdvBench & Harmful instruction benchmark for evaluating jailbreak success & Chen et al.~\cite{Chen2024-lv}, Wang et al.~\cite{Wang2024-zh}, Mangaokar et al.~\cite{Mangaokar2024-ii} & One of the most common text-based jailbreak benchmarks. \\

HarmBench & Standardized evaluation framework for harmful behaviors and jailbreak attacks & Chen et al.~\cite{Chen2024-lv}, Li et al.~\cite{Li2024-pj} & Widely used for stronger and more standardized jailbreak evaluation. \\

JailbreakBench & Open robustness benchmark with standardized behaviors, prompts, and evaluation framework & Li et al.~\cite{Li2024-pj} & Often used in follow-up work to compare jailbreak attacks under a shared evaluation setup. \\

Behaviors dataset (DRA) & Harmful behavior set used in disguise-and-reconstruction style jailbreak evaluation & Liu et al.~\cite{Liu2024-rd} & A task-specific harmful-behavior set used in DRA-style attack evaluation. \\

SafeBench & Multimodal safety benchmark for LVLM jailbreak and harmful instruction following & Gong et al.~\cite{Gong2023-xr} & Introduced for evaluating cross-modal safety risks in VLMs. \\

\bottomrule
\end{tabular}
\label{tab:sec4-datasets}
\end{table*}

Overall, studies on unintentional harmful generation often rely on broad trustworthiness benchmarks such as \textit{TrustGPT} and \textit{DecodingTrust}, whereas intentional attack papers increasingly converge on shared jailbreak benchmarks such as \textit{AdvBench}, \textit{HarmBench}, and \textit{JailbreakBench}. For multimodal attacks, benchmark construction is still evolving, with datasets such as \textit{SafeBench} emerging to evaluate cross-modal safety vulnerabilities.

\color{black}

\section{LLMs as Guardians: Potential (RQ3)}
\label{sec:RQ3} 

While this survey primarily focuses on categorizing methods and approaches, we observe several general performance trends reported across prior work. For example, fine-tuned classifiers (e.g., RoBERTa-based models) consistently demonstrate strong performance on standard benchmark datasets such as Jigsaw. In contrast, LLMs show competitive zero-shot or few-shot capabilities, particularly in handling implicit or context-dependent harmful content, although their performance varies across tasks and evaluation settings. Despite efforts to improve robustness to adversarial attacks such as jailbreaks, existing defenses remain limited, with many studies highlighting persistent vulnerabilities. These observations reflect both the progress made and the remaining challenges in LLM safety.

With LLMs' advanced language understanding, they are increasingly recognized for their potential. We organize this section based on the tasks they perform. First, they are used to build datasets \citep{das2024offensivelang, 392, 383} or generate counter-speech \citep{hong-etal-2024-outcome,105,263,422}. Second, LLMs' reasoning capability can help classify diverse forms of harmful content \citep{masud2023focal, nasir2023llms, dutta2023modeling,xu2024hatespeech, Agarwal2023-ml, ovalle2023you} and perform content moderation \citep{lee2024exploring, wang2023evaluating, he2024you, vishwamitra2024moderating, lu2024toxic, wang2023self, ovalle2023you, zhang2024genderalign}. Finally, previous research efforts show that LLMs can contribute to the prevention ~\citep{xie2023empirical,ji2023beavertails,lee2023square,bang2023enabling,qureshi2024refine,kaneko2024evaluating, wang2024model,jorgensen2023improving} and the defense against jailbreaking threats \cite{Phute2023-ue, Chen2023-eq, kumar2023certifying, beck25sensitivity}.


Then, we explore the potential of LLMs in two areas: addressing existing harmful content in the first three subsections, and preventing potentially harmful generation in the latter two subsections, respectively. Cross several key tasks, namely mitigating existing harmful content through generation (Section~\ref{Text generation-Potentials}), identifying and classifying harmful content (Section~\ref{Text Classification-Potentials}), content moderation (Section~\ref{Moderation-Potentials}), and prevention (Section \ref{Prevention-Potentials}), while simultaneously examining crucial strategies for the defense against jailbreak and the prevention of LLM-related risks (Section~\ref{Jailbreak-Potentials}).


\subsection{Mitigating Existing Harmful Content through Generation}
\label{Text generation-Potentials} 

Researchers have leveraged LLMs to generate synthetic harmful content datasets, which enable the development of more effective detection/classification models and reduce reliance on ethically sensitive real-world data. LLMs also contribute to counter-speech generation, where LLM responses effectively combat online harmful content in real-time. Additionally, LLMs can be used for red-teaming other LLMs, where controlled harmful content generation helps stress-test moderation systems and expose vulnerabilities in content filtering mechanisms. The potential of LLMs' text generation in harmful content is thus demonstrated from these three aspects.

\subsubsection{Harmful Content Generation for Dataset Construction} 
\label{dataset construction} \leavevmode\newline
LLMs have been increasingly used to generate synthetic harmful content datasets, providing valuable resources for harmful content detection, classifier evaluation, and safety alignment. These datasets differ in language coverage, dataset construction methods, and target applications, ensuring a comprehensive and diverse approach to detecting and mitigating toxic content.


Several studies focus on generating explicit and implicit harmful content datasets using LLMs. For example, OffensiveLang contains 8,270 ChatGPT-generated offensive and non-offensive texts spanning 38 target groups across race, gender, religion, and disability, designed to enhance classifier robustness against implicit toxicity \cite{das2024offensivelang}. Similarly, K-OMG produces culturally aligned offensive language datasets for Korean. It leverages prompt-based LLM generation to address low-resource language detection challenges \cite{392}. ParaFusion generates 3.5 million paraphrases and filters offensive content to improve lexical and syntactic diversity in harmful content detection to ensure a balanced dataset for NLP applications \cite{383}.

Datasets targeting bias and stereotypes are also constructed using LLMs. SeeGULL Multilingual introduces 25,861 LLM-generated stereotypes across 20 languages and provides a benchmark for evaluating fairness and bias mitigation in harmful content detection \cite{397}. Meanwhile, a GPT-generated dataset annotated for gender bias quantifies bias degrees in offensive text, offering fine-grained insights into identity-based attacks and biased language patterns \cite{286}. 

LLM-generated datasets cover various forms of harmful content, from explicit slurs to coded language and microaggressions. Some datasets focus on identity-based harmful content, targeting racial, gender, and religious biases \cite{das2024offensivelang}, while others address sarcasm and indirect harmful content. KoCoSa, a Korean Context-aware Sarcasm Detection Dataset, integrates LLM-generated sarcastic dialogues to train harmful content detection models on contextual toxicity \cite{kim-etal-2024-kocosa}. Certain datasets focus on multilingual and cultural variations in harmful content. IndicLLMSuite provides LLM-generated toxic prompts across 22 Indic languages, enabling multilingual harmful content detection \cite{394}. Similarly, SeeGULL Multilingual captures geo-culturally situated stereotypes, which offers insights into harmful content variations across different regions \cite{397}.

Harmful content datasets generated by LLMs are widely applied in benchmarking detection models and stress-testing moderation systems. GPT-HateCheck generates functional harmful content test cases. These test cases address the limitations of template-based benchmarks like HateCheck and ensure more nuanced and real-world harmful content samples \cite{hatecheck}. Counterfactually Augmented Data (CAD) expands harmful content datasets with adversarial variations, improving classifier generalization across different harmful content contexts \cite{283}. CAD-based dataset augmentation has been explored for security-related text classification. It has shown potential in harmful content detection, spam detection, and fraud detection \cite{223}.

Last, LLM-generated datasets also play a role in safety alignment and ethical model training. KoTox, a Korean Toxic Instruction Dataset, contains 39,000 toxic instruction-output pairs to help LLMs reject harmful queries, strengthening ethical AI alignment \cite{264}. Additionally, IndicLLMSuite generates toxic prompts and non-toxic responses for Indic languages, contributing to harmful content detection efforts in underrepresented linguistic contexts \cite{394}.

\subsubsection{LLMs for Counter Speech Generation}
\label{counterspeech} \leavevmode\newline
Counter speech has emerged as a key strategy for mitigating online harmful content, and recent advances in LLMs have opened new possibilities for automating its generation. Researchers have explored various approaches to improve the effectiveness of LLM-generated counter-speech. They focus on aspects such as conversation outcome constraints \cite{hong-etal-2024-outcome}, discourse structuring \cite{105}, instructional prompting \cite{263}, and zero-shot performance analysis \cite{422}. 

Hong et al. (2024) propose to enhance LLM-generated counter speech by optimizing for conversation outcomes. It uses reinforcement learning and instruction prompting to encourage civil engagement and reduce incivility in social media discussions \cite{hong-etal-2024-outcome}. Another approach proposed by Hedderich et al. (2024) applies LLM-powered chatbots for cyberbullying intervention. Thus, teachers are allowed to design interactive training tools where students practice counter-speech in a safe, role-playing environment \cite{105}.

Beyond these applications, a discourse-aware framework (DisCGen) improves counter-speech generation by incorporating Segmented Discourse Representation Theory (SDRT), which leads to more diverse and contextually relevant responses \cite{263}. A zero-shot evaluation comparing GPT-2, DialoGPT, ChatGPT, and FlanT5 further demonstrates that ChatGPT produces the most effective counter-speech, but larger models risk generating more toxic responses. The study by Saha et al. (2024) also finds that manual prompting outperforms other strategies, though LLMs struggle with humor and rhetorical counter-speech \cite{422}.

\subsubsection{Harmful Content Generation for Red-Teaming}
\label{red-teaming} \leavevmode\newline
Red-teaming LLMs plays a critical role in identifying vulnerabilities and evaluating safety mechanisms, particularly in the context of harmful content generation. By systematically probing models with adversarial prompts, researchers can stress-test moderation systems and enhance content safety mechanisms. Recent studies have explored a variety of automated red-teaming techniques leveraging LLM text generation, ranging from reinforcement learning-based attacks to optimization-driven adversarial auditing.

One approach is reinforcement learning-based adversarial prompt generation, as demonstrated by the three-stage red-teaming framework (Exploration, Establishment, and Exploitation), which systematically manipulates LLMs to increase their rate of toxic output, exposing hidden vulnerabilities that standard safety techniques fail to address \cite{27}. Another strategy is discrete optimization-based auditing, where frameworks like ARCA use log-probability approximations to search for high-risk input-output pairs, making them more effective than manual red-teaming methods \cite{123}.

More recent methods leverage LLM-based adversarial auditing. TroubleLLM introduces text style transfer and adversarial conditioning techniques. It generates highly effective adversarial prompts that outperform human-crafted and template-based red-teaming methods, which makes it an automated solution for systematically testing model vulnerabilities \cite{385}. Similarly, GFlowNet fine-tuning with MLE smoothing enables the creation of diverse and transferable attack prompts. The generated adversarial testing is more robust and adaptable compared to traditional reinforcement learning-based strategies \cite{520}.

Across the three generative uses, one pattern separates effective contributions from weak ones: value depends on validation against the harm of interest, not on output volume. In dataset construction, the studies that advance the field target a specific gap, as OffensiveLang and K-OMG extend coverage to implicit and low-resource offensive language \cite{das2024offensivelang, 392}, SeeGULL Multilingual and the gender-bias corpora isolate identity-based harm for fairness evaluation \cite{397, 286}, and GPT-HateCheck builds functional test cases that template benchmarks miss \cite{hatecheck}. Their shared weakness is that synthetic text inherits the generator's own biases, so a dataset's usefulness rests on measured downstream gains rather than on its size. Counter-speech and red-teaming show the same dependence on the right objective: outcome-constrained and discourse-aware generation outperform unconstrained prompting \cite{hong-etal-2024-outcome, 263}, larger models produce more fluent replies but also more toxic ones \cite{422}, and optimization- or LLM-driven red-teaming now exceeds human-written prompts in success rate and diversity \cite{385, 520}, which is why these generators double as the evaluation tools of Section~\ref{Jailbreak-Potentials}. To conclude, the generation methods combating the harmful content, they are used to broaden the dataset, create information against harmful speeches, and red-teaming the existing system to look for potential vulnerabilities. As for the dataset building, the trend is from monolingual, explicit-toxicity sets toward multilingual, implicit, and culturally situated data. The other recurring trade-off is fidelity versus contamination: synthetic data scales coverage but can encode the generator's own biases, so its value depends on downstream validation rather than volume alone.

\subsection{Identifying and Classifying Harmful Content}
\label{Text Classification-Potentials}

\begin{table}[b]
\centering
\caption{Summary of major LLM-based text classification strategies for harmful-content detection.}
\label{tab:techniques_summary_hierarchical_models}
\footnotesize      
\begin{tabular}{@{}p{2cm}p{2.8cm}p{5.2cm}p{1.6cm}@{}}
\toprule
\textbf{Strategy Type} & \textbf{Model} & \textbf{Key Techniques} & \textbf{Papers} \\
\midrule

\multirow{3}{=}{Training-Based}  
  & BERT, RoBERTa, LLaMA          & Supervised adaptation on harmful content datasets; domain-specific optimization                           & \citep{masud2023focal, nasir2023llms, dutta2023modeling} \\ 
  & GPT-3.5, GPT-4 & RLHF for ethical alignment                                & \citep{li2025more} \\  
  & LLaMA, Mistral                & Robust fine-tuning with adversarial examples to resist prompt attacks                                  & \citep{zhou2024multimodal, wu2024exploring} \\ 
\midrule
\addlinespace
\multirow{2}{=}{Prompt-Based}  
  & GPT-3, LLaMA         & Zero-shot / few-shot learning using task-specific prompt templates                                      & \citep{xu2024hatespeech, Agarwal2023-ml, ovalle2023you} \\ 
  &  GPT-4                & Output filtering or reformulation of toxic generations (e.g., via contrastive decoding or classifiers)   & \citep{Zhang2024, cao2024toxicity} \\ 
\midrule
\addlinespace
\multirow{1}{=}{Bias Mitigation}  
  & BERT, LLaMA                   & Counterfactual data augmentation, debiasing objectives, demographic analysis                             & \citep{You2024, Liu2024, sun2024bias} \\
\midrule
\addlinespace
\multirow{1}{=}{Data-Centric}  
  & GPT-3.5, GPT-4                & LLM-generated harmful content and counter-speech data augmentation                                            & \citep{meguellati2025llm} \\ 
\midrule
\addlinespace
\multirow{1}{=}{Cross-Lingual \& Context}  
  & mBERT, XLM-R, LLaMA           & Multi-language model adaptation for low-resource harmful content classification                                 & \citep{bigoulaeva2021cross, ghorbanpour2025data} \\ 
  & RoBERTa, LLaMA                & Coreference resolution and contextual history modeling for implicit hate                                   & \citep{xu2024hatespeech} \\
\midrule
\addlinespace
Evaluation-Based & All models                     & Perplexity-based scoring, calibration curves, benchmark comparison                                          & \citep{jansen2022perplexed, lim2023evaluating} \\

\bottomrule
\end{tabular}
\end{table}

Despite significant progress in text recognition capabilities of LLM in recent years, harmful content remains a complex and evolving challenge in cyberspace. Currently, the main method for detecting such harmful content is text classification, which assigns structured labels such as "harmful", "harmless", "hateful", "insulting", etc. to unstructured text. Owing to their strong contextual reasoning and ability to generalize from examples, LLMs have demonstrated exceptional performance in this task. Nevertheless, these models also introduce new concerns, including the amplification of pre-existing social biases, difficulty in detecting subtle or implicit forms of hate, and vulnerability to adversarial manipulation.

This section presents a structured examination of text classification methods for harmful content detection. We begin by examining how fine-tuning techniques enhance model adaptability and detection accuracy. It then discusses the role of diverse model architectures in improving contextual understanding and multilingual detection capabilities. In addition, we explore prompt engineering strategies to reduce biases and annotation costs while improving interpretability. To address dataset scarcity and class imbalance, we discuss data augmentation and synthetic text synthesis. Cross-lingual transfer learning leverages multilingual transformers to extend detection to low-resource languages. Subsequently, we analyze the contributions of diverse model architectures to enhance contextual sensitivity and enable robust performance in multilingual settings. 

\subsubsection{Characterizing Toxicity} 
\leavevmode\newline
Defining and measuring toxicity present inherent challenges. Luo et al. \cite{luo2023legally} highlighted a potential misalignment between LLM-generated harmful content and established legal definitions, pointing to inconsistencies with regulatory goals. Furthermore, the perception and manifestation of harmful content can vary culturally. Lee et al. \cite{lee2024exploring} demonstrated this by collecting datasets showing variability in toxicity norms across different English-speaking regions, including North America, Australia, Singapore, and the United Kingdom. Evaluating the trustworthiness of LLMs requires a multi-faceted approach. Qian et al. \cite{qian2024towards} proposed evaluating trustworthiness across five dimensions—reliability, toxicity, privacy, fairness, and robustness—using linear probing. Their findings suggest that a model's capacity for stable and secure generation can often be discerned even in early pretraining checkpoints.

\subsubsection{Prompt Engineering}
\leavevmode\newline
Prompt engineering has become instrumental for enhancing model transparency, interpretability, and fairness in harmful content detection. Carefully crafted prompts explicitly guide model reasoning, significantly reducing annotation efforts and improving accuracy and ethical compliance by explicitly instructing models to avoid stereotypical biases. For example, structured prompt templates have been shown to reduce the generation of harmful stereotypes \citep{dorner2022human, rosenblatt2022critical}. Identity aware prompting mitigates fairness-conditioned prompts improve safety compliance \cite{Agarwal2023-ml, ovalle2023you}. Context-sensitive prompts have been shown to significantly enhance transparency and alignment with human values \cite{Som2023-yq}. Moreover, advanced prompting methods, including zero shot and few shot learning, further facilitate generalization to novel harmful content patterns and implicitly toxic content, enhancing model robustness and reducing potential biases in real-world deployments \citep{xu2024hatespeech, Zhang2024}.

\subsubsection{Fine-tuning and Alignment for Harmful Content Detection}
\leavevmode\newline
Fine-tuning LLMs remains one of the most effective strategies for domain adaptation in harmful content detection. By adjusting pretrained models to task-specific objectives and sensitive content distributions, fine-tuning enhances recognition of implicit toxicity, subtle insults, and identity-directed hate. For instance, label-sensitive fine-tuning explicitly exposes underrepresented hate cases, improving performance on rare but critical examples \cite{tuck2024unmasking}, while staged pipelines progressively capture nuanced forms of aggression across diverse datasets \cite{kumar2024towards}. To address class imbalance, a focal loss strategy reweights examples based on classification difficulty, yielding substantial gains on toxic content benchmarks \cite{masud2023focal}, and curriculum learning further boosts robustness by gradually exposing the model to increasingly challenging toxic inputs \cite{nasir2023llms}.

In multi-label settings, frameworks that leverage inter-label correlations and hierarchical dependencies enable finer-grained detection of overlapping toxic behaviors \cite{dutta2023modeling}. To mitigate fairness concerns and domain drift, demographic-sensitive training and episodic fine-tuning tailor models to underrepresented groups and evolving content distributions \cite{hickey2025peripatetic}, while experiments on Arabic toxic corpora highlight ongoing challenges in multilingual generalization \cite{almohaimeed2023thos}. Generative replay mechanisms have been introduced during fine-tuning to retain ethical constraints and prevent catastrophic forgetting \cite{pendzel2024generative}. Finally, hybrid approaches that combine fine-tuning with prompt engineering, such as prompt-based supervision to resist adversarial jailbreaks \cite{zhou2024multimodal} and dynamic prompting to guide models through ambiguous or subversive inputs \cite{Gupta2024} alongside reconfiguration techniques for identifying harmful intent and suppressing undesirable outputs \cite{wu2024exploring}, further enhance both robustness and ethical alignment.

\subsubsection{Data Augmentation and Text Synthesis for the Imbalance in Speech Categories}
\leavevmode\newline
Data augmentation techniques and synthetic data generation address dataset scarcity and class imbalance in harmful content detection. Back-translation and paraphrasing via pretrained models can produce diverse rephrasings of toxic and non-toxic content, effectively expanding training distributions \citep{kolla2022study}. Prompt-based synthesis with large generative models has been shown to generate realistic hate and non-hate examples, improving robustness to novel linguistic patterns \citep{maity-etal-2024-toxvidlm}. Embedding-based matching further refines augmented samples to maintain semantic coherence, boosting generalization across varied expressions of toxicity \citep{hossain2024align}.

\subsubsection{Cross-lingual Transferability for Classification}
\leavevmode\newline
Cross-lingual transfer learning extends detection models to resource-scarce languages by leveraging shared multilingual representations. Multilingual transformers fine-tuned on high-resource datasets can be adapted to low-resource contexts with minimal labeled data, yielding significant performance gains \citep{bigoulaeva2021cross, ghorbanpour2025data}. 
Coreference-driven approaches explicitly model referential chains (pronouns, named entities) within text, capturing subtle, identity-targeted hate that relies on context and anaphora. 

\subsubsection{Evaluation and Benchmarking for Robustness and Fairness}
\leavevmode\newline
Rigorous evaluation and benchmarking methodologies ensure reliable, fair, and transparent performance in harmful content detection models. Benchmarking studies systematically evaluate models across diverse harmful content datasets, explicitly addressing issues of bias, fairness, and generalization capabilities \citep{vogel2023explaining}. Comprehensive surveys guide the selection and development of robust metrics and datasets, fostering improved model reliability and fairness across contexts \citep{lim2023evaluating}. Furthermore, specialized benchmarking datasets explicitly tailored for harmful content enhance robustness evaluations, accommodating diverse linguistic variations and nuanced offensive language patterns \citep{cao2024toxicity}. Joint dataset creation and benchmarking initiatives facilitate consistent model comparisons and accelerate knowledge transfer, improving harmful content detection performance across research communities \citep{hossain2024misgendermender}. Additionally, established safety taxonomies systematically categorize prompt-related risks, further supporting the safe and ethical deployment of models \citep{66}.

Within detection, the technique families trade off along a consistent axis of supervision cost against adaptability. Training-based methods give the highest in-distribution accuracy but need labeled data and degrade under distribution shift, which is why the imbalance-focused variants differ mainly in how they reweight rare cases: focal loss adjusts per-example difficulty \cite{masud2023focal}, curriculum learning orders examples from easy to hard \cite{nasir2023llms}, and multi-label frameworks share signal across co-occurring harms \cite{dutta2023modeling}. Prompt-based methods invert the trade-off, cutting annotation cost and generalizing to novel and implicit patterns \cite{xu2024hatespeech, Zhang2024} but proving less reliable under adversarial input. Data augmentation and cross-lingual transfer act as complements rather than alternatives, since back-translation, paraphrase, and embedding-matched synthesis expand scarce training distributions \cite{kolla2022study, maity-etal-2024-toxvidlm, hossain2024align} while multilingual transformers carry detection into low-resource languages \cite{bigoulaeva2021cross, ghorbanpour2025data}, each feeding the training- and prompt-based methods rather than replacing them.

\subsection{Content Moderation} 
\label{Moderation-Potentials} 

The rapid proliferation of LLMs, particularly those built upon Transformer-based architectures, has completely changed the application of natural language processing. However, with these advances, there are growing concerns that LLMs may generate harmful content, either inadvertently or through adversarial prompting \cite{zhang2023joint}. As a result, ensuring the safe and responsible deployment of LLMs has become a pressing research priority, with content moderation emerging as a critical area of focus. Recent advancements demonstrate that with careful prompt design, fine-tuning, or parameter editing, LLMs can perform moderation at scale while preserving language diversity and generation fluency. This subsection surveys technical approaches that transform LLMs into effective content moderators across monolingual and multilingual environments.

\subsubsection{Prompt Engineering} 
\leavevmode\newline
Prompt engineering offers a means of controlling LLM outputs. Wang et al. \cite{wang2023evaluating} used GPT-3 itself, prompted appropriately, to generate explanations for online content moderation decisions. He et al. \cite{he2024you} explored detoxification in GPT-2 and T5 using prompt tuning. Vishwamitra et al. \cite{vishwamitra2024moderating} developed a reasoning-based Chain-of-Thought (CoT) prompting technique that automatically generates and updates prompts for moderation purposes. A more direct approach involves vocabulary manipulation; Lu et al. \cite{lu2024toxic} proposed pruning subword tokens associated with toxic words within trained LLMs. This simple mechanism demonstrated improved dialogue diversity and toxicity reduction in models like Llama-3.1\_8B. 

\subsubsection{Alignment, and Parameter Editing} 
\leavevmode\newline
Researchers are actively developing methods to mitigate these risks, broadly categorized into safety training and safeguards, often employed concurrently \cite{wang2023self}. Several approaches focus on refining the models during or after training. Fine-tuning has proven effective in moderating specific biases, such as those related to pronouns or race \cite{ovalle2023you}. Alignment techniques aim to steer model behavior towards desired norms. Zhang et al. \cite{zhang2024genderalign} generated a specialized alignment dataset to mitigate gender bias in LLMs specifically. Similarly, Direct Preference Optimization (DPO) trained on English preference data has shown promise in reducing toxicity even in multilingual generation across models like mGPT-1.3B, BLOOM, Llama3, and Aya-23 \cite{li2024preference}. Contrastive adversarial gender debiasing leverages contrastive learning techniques, reportedly outperforming standard adversarial debiasing methods \cite{torres2024contrastive}.

Beyond extensive retraining or fine-tuning, modifying specific model parameters offers a more targeted approach. Wang et al. \cite{Wang2024-ac} discovered that editing only a small subset of model parameters can effectively modulate and detoxify generation while preserving the LLM's general capabilities. Building on this, Elesedy et al. \cite{elesedy2024lora} employed low-rank adapters (LoRA), demonstrating effective content moderation with significantly fewer (100x) guardrail parameters compared to full fine-tuning, preventing degradation of the generative model and suggesting potential for moderated models on resource-constrained devices like mobile phones \cite{elesedy2024lora}. 

Unlearning techniques aim to selectively remove specific knowledge or behaviors, such as copyrighted or harmful content, from a trained model without degrading its overall utility. Negative Preference Optimization (NPO) is recognized as a state-of-the-art unlearning method. Fan et al. \cite{fan2024simplicity} introduced a simplified preference optimization method for unlearning that removes the need for a reference model, potentially making the process more efficient. 

\subsubsection{Cross-Lingual Considerations}
\leavevmode\newline
Toxicity and its mitigation are not confined to a single language. Interestingly, safety alignment efforts can exhibit cross-lingual transfer. Li et al. \cite{li2024preference} found that training in DPO solely with English data could reduce toxicity in multilingual outputs. Lu et al. \cite{lu2024every} further corroborated this multilingual potential, showing that unsafe generation patterns can influence outputs across languages. Crucially, their experiments indicated that unlearning methods applied by addressing harmful responses in both English and the content's original language could effectively eliminate toxic generation across multiple languages supported by the model.

In summary, addressing toxicity in LLMs involves understanding its nuanced nature across contexts and languages and developing a diverse toolkit of mitigation strategies, ranging from comprehensive retraining and alignment to targeted parameter editing, sophisticated prompt engineering, and selective unlearning techniques. The moderation methods are divided by how invasively they alter the model. Prompt-based moderation is the least invasive, steering a frozen model through chain-of-thought reasoning or decision explanations \cite{vishwamitra2024moderating, wang2023evaluating} and even simple toxic-subword pruning \cite{lu2024toxic}, but it gives the weakest guarantees against determined evasion. Alignment and parameter editing intervene more deeply and more durably, as targeted editing and contrastive debiasing reshape specific behaviors \cite{Wang2024-ac, torres2024contrastive} while parameter-efficient guardrails and unlearning reach similar effect at lower cost and with less capability loss \cite{elesedy2024lora, fan2024simplicity}. The choice is therefore one of permanence against flexibility, with prompt-based control suited to fast-changing policies and parameter-level methods to stable, high-volume deployment.

\subsection{Potential in Harm Prevention}
\label{Prevention-Potentials} 

Beyond filtering or correcting harmful content after generation, prevention steers models toward safe outputs by design. Several techniques relevant here, including fine-tuning and alignment, parameter editing, and data augmentation, were already covered as detection and moderation methods in Sections~\ref{Text Classification-Potentials} and~\ref{Moderation-Potentials}. To avoid redundancy, we focus on what is specific to prevention, namely anticipatory control exercised at the data, prompt, and representation levels before any harmful completion is produced. At the data level, counterfactual augmentation \cite{xie2023empirical} and preference supervision such as Beavertails, SQuARe, value-aligned classifiers, and REFINE-LM \cite{ji2023beavertails, lee2023square, bang2023enabling, qureshi2024refine} shape behavior before deployment. At the prompt level, chain-of-thought reduces unconscious social bias \cite{kaneko2024evaluating}. At the representation and decoding level, prevention becomes proactive through parameter editing for robustness \cite{wang2024model}, mean-centred activation steering \cite{jorgensen2023improving}, and controllable generation such as Gated Toxicity Avoidance \cite{kim2023gta}. A recurring tension is that safety tuning can degrade utility, which ForgetFilter counters by preventing forgetting \cite{zhao2024learning}, SteerLM addresses by granting inference-time control without RLHF \cite{dong2023steerlm}, and Antidote handles by pruning harmful parameters introduced through fine-tuning attacks \cite{huang2024antidote}, with self-supervised monitoring complementing human labels \cite{jain2023bring}.

At the data level, prevention enriches the supervision a model receives before deployment. Xie et al.~\cite{xie2023empirical} employ counterfactual data augmentation (CDA) as a bias mitigation strategy for pretrained language models, specifically GPT-2 and BERT.  Beavertails \cite{ji2023beavertails} constructs a comprehensive Human-Preference Dataset comprising 333,963 question–answer pairs and 361,903 expert-labeled comparison instances, annotated for helpfulness and harmlessness. This dataset contributes significantly to enhancing LLM safety through RLHF. Lee et al. \cite{lee2023square} build a Korean language dataset of sensitive questions with corresponding acceptable and unacceptable responses. And the experiments show the dataset improves the acceptable responses for HyperCLOVA \cite{yoo2024hyperclova} and GPT-3 \cite{brown2020language}. Some additional components are helpful for prevention. Bang et al. \cite{bang2023enabling} introduce an additional fine-tuned classifier performing value-aligned supervision, suggesting that the value-based classifier improves inclusivity and explainability in LLM. REFINE-LM \cite{qureshi2024refine} involves a simple model added on top of LLMs, which is cheaper to train with reinforcement learning and reduces biases while preserving the performance.  

At the prompt level, structured reasoning can suppress harm before it surfaces, as chain-of-thought prompting reduces the unconscious social bias a model expresses during step-by-step generation \cite{kaneko2024evaluating}. The representation and decoding levels are where prevention becomes genuinely proactive, because they act on the model's internal computation rather than its inputs. Early methods intervene globally: parameter editing hardens a model against adversarial signals by adjusting a small set of weights \cite{wang2024model}, and activation steering derives control vectors by subtracting the mean activation of a target dataset from the model's activations to steer generation away from toxicity at inference time \cite{jorgensen2023improving}. Recent work makes these interventions selective. Conditional Activation Steering learns a condition vector that detects whether an input falls in a target category and applies the refusal direction only then, so the model refuses harmful prompts while still answering benign ones \cite{lee2025programming}, and ToxEdit applies the same selective principle at the parameter level, detecting toxic activation patterns during the forward pass and routing computation through anti-toxic pathways to avoid the over-editing that makes earlier knowledge-editing methods reject legitimate queries \cite{lu2025adaptive}. At the decoding level, controllable generation has advanced from gated avoidance of toxic topics, tones, and phrasings \cite{kim2023gta} to discrete auto-regressive biasing, which steers token selection toward detoxified output while preserving fluency at low decoding cost \cite{pynadath2025controlled}.

Prevention repeatedly meets a tension between safety and utility, and several methods are defined by how they manage it. Safety fine-tuning can degrade downstream performance by causing a model to forget useful data, which ForgetFilter addresses as a pre-finetuning step that preserves capability while improving safety \cite{zhao2024learning}. SteerLM uses supervised fine-tuning to give users control over model behavior at inference time while staying competitive with RLHF-based systems \cite{dong2023steerlm}, and Antidote defends against fine-tuning attacks by detecting and pruning the parameters they introduce, blocking harmful generation without sacrificing performance \cite{huang2024antidote}. Self-supervised evaluation complements human labels by monitoring behavior continuously rather than only at fixed checkpoints \cite{jain2023bring}. These methods share the aim we revisit in Section~\ref{sec:RQ3}: fine-grained, context-sensitive control that raises safety without the over-refusal and reduced diversity that blanket safety tuning tends to produce.

\subsection{Defense against LLM Jailbreaking}
\label{Jailbreak-Potentials} 

The current landscape of LLM jailbreak research has witnessed the development of increasingly sophisticated attack and defense mechanisms. These approaches range from fundamental text-based manipulations to advanced multimodal strategies, reflecting the evolving complexity of adversarial threats against LLMs. Defense mechanisms have similarly evolved, incorporating adaptive self-monitoring, efficient safeguards, and context-aware strategies to enhance model security. This technical progression demonstrates both the increasing sophistication of attack methodologies and the corresponding advancement in protective measures, highlighting the ongoing challenge of maintaining LLM safety while preserving model utility.

\subsubsection{Adaptive and Self-monitoring Defenses} 
\leavevmode\newline
Adaptive defense mechanisms leverage LLMs' internal monitoring capabilities and systematic evaluation frameworks to provide real-time and certified protection. Phute et al. (2023)~\cite{Phute2023-ue} introduce the LLM Self Defense approach, enabling models to screen their own outputs for harmful content without additional fine-tuning. Chen et al. (2023)~\cite{Chen2023-eq} propose Moving Target Defense strategies, employing randomized selection from multiple LLM instances to minimize attack success rates, reducing them from 37.5\% to 0\%. Furthermore, Kumar et al. (2023)~\cite{kumar2023certifying} develop the erase-and-check framework, offering certified safety guarantees against adversarial prompting with 92\% detection accuracy. Wang et al. (2023)~\cite{Wang2023-vh} and Rao et al. (2023)~\cite{Rao2023-pr} provide comprehensive frameworks for systematic safety evaluation, using detailed taxonomies and standardized testing to assess potential vulnerabilities effectively.

\subsubsection{Scalable Safeguards and Context-aware Defenses} 
\leavevmode\newline
Efficient defense mechanisms aim to minimize computational overhead while maintaining robust security. Kim et al. (2023)~\cite{Kim2023-mj} develop the Adversarial Prompt Shield (APS), a lightweight classifier using only 25\% of the parameters compared to traditional solutions, achieving state-of-the-art performance. Additionally, these efficient defenses align with broader evaluation frameworks that prioritize both computational efficiency and model robustness, ensuring that security measures do not compromise processing speed or system scalability.
Developing effective defenses requires addressing social and contextual factors that influence model behavior. Sociodemographic prompting \cite{beck25sensitivity} is a finding that motivates context-aware defenses: model responses, and therefore the safeguards they require, vary with the demographic framing of a prompt, which argues for socially informed rather than uniform protection.

\subsubsection{Evaluation and Mechanistic Analysis for Jailbreak Defense}
\leavevmode\newline
Effective evaluation and mechanistic analysis are critical for understanding LLM vulnerabilities and informing robust jailbreak defenses. In terms of safety evaluation, Wang et al. (2023)~\cite{Wang2023-vh} introduce the Do-Not-Answer dataset, offering a comprehensive benchmark for evaluating model responses to high-risk prompts across various scenarios. This dataset provides a hierarchical risk taxonomy covering multiple harm categories, enabling systematic assessments of LLM safety. Building on this foundation, Yu et al. (2024)~\cite{Yu2024-pm} develop a holistic framework that characterizes and evaluates LLM reliability across diverse jailbreak strategies, incorporating key metrics such as attack success rates, toxicity scores, and grammatical accuracy to provide a multi-dimensional view of LLM robustness. These evaluation frameworks have been further enhanced by recent works such as Chen et al. (2024)~\cite{Chen2024-lv}, Wang et al. (2024)~\cite{Wang2024-zh}, and Mangaokar et al. (2024)~\cite{Mangaokar2024-ii}, which provide additional metrics and benchmarks for assessing different attack strategies. Complementing these evaluation frameworks, mechanism analysis has revealed deeper insights into LLM vulnerabilities. Zhou et al. (2024)~\cite{Zhou2024-uh} conduct a detailed analysis of hidden states in LLMs, demonstrating how jailbreak prompts circumvent safety mechanisms by disrupting intermediate representations rather than directly attacking initial ethical recognition. This understanding has been enriched by studies from Jiang et al. (2023)~\cite{Jiang2023-ee}, Liu et al. (2024)~\cite{Liu2024-rd}, Li et al. (2024)~\cite{Li2024-pj}, and Lin et al. (2024)~\cite{Lin2024-ly}, which explore various compositional attack mechanisms. Further theoretical work by Shen et al. (2024)~\cite{Shen2024-fe} in \textit{Mission Impossible} proves that adversarial vulnerabilities are inevitable under current training paradigms, while research from Gong et al. (2023)~\cite{Gong2023-xr} and Shayegani et al. (2023)~\cite{Shayegani2023-in} extends these insights to multimodal contexts. The defensive aspects of these mechanisms have been explored by Phute et al. (2023)~\cite{Phute2023-ue}, Chen et al. (2023)~\cite{Chen2023-eq}, Kim et al. (2023)~\cite{Kim2023-mj}, and Kumar et al. (2023)~\cite{kumar2023certifying}, highlighting the importance of comprehensive assessment approaches that consider both empirical performance and theoretical limitations. These combined insights from both evaluation frameworks and mechanism analyses suggest the need for new training strategies to enhance robustness against adversarial prompting.

\begin{table}[b]
  \caption{Technical classification of representative jailbreak attacks and defense mechanisms.}
    \label{tab:jailbreak-technical}
  \footnotesize
  \centering
  \begin{tabular}{lp{2cm}p{3.5cm}>{\raggedright\arraybackslash}p{3cm}}
    \toprule
    \textbf{Technical Approach} & \textbf{Methods} & \textbf{Key Techniques} & \textbf{Papers} \\
    \midrule
    \textbf{Reinforcement Learning-based} & RL-JACK & RL-based prompt optimization \newline
    Black-box attack framework \newline
    Dynamic refinement strategies & Chen et al.(2024)\cite{Chen2024-lv} \newline
    Wang et al.(2024)\cite{Wang2024-zh} \newline
    Kumar et al.(2023)\cite{kumar2023certifying} \\
    \midrule
    \textbf{Gradient Optimization} & ASETF \newline
    PGD \newline
    GradAttack & Adversarial suffix embeddings \newline
    Gradient-based optimization \newline
    Low computational cost \newline
    Token-level perturbation & Wang et al.(2024)\cite{Wang2024-zh} \newline
    Mangaokar et al.(2024)\cite{Mangaokar2024-ii} \newline
    Li et al.(2024)\cite{Li2024-pj} \\
    \midrule
    \textbf{Universal Adversarial} & PRP \newline
    UAP \newline
    AdvPrefix & Universal adversarial prefixes \newline
    Cross-architecture attack \newline
    Guard mechanism bypass \newline
    Transfer learning attack & Mangaokar et al.(2024)\cite{Mangaokar2024-ii} \newline
    Jiang et al.(2023)\cite{Jiang2023-ee} \newline
    Liu et al.(2024)\cite{Liu2024-rd} \\
    \midrule
    \textbf{Compositional Attacks} & Prompt Packer \newline
    DRA \newline
    DrAttack \newline
    ABJ & Complex task embedding \newline
    Instruction disguising \newline
    Sub-prompt decomposition \newline
    Task-based hiding \newline
    Semantic camouflage & Jiang et al.(2023)\cite{Jiang2023-ee} \newline
    Liu et al.(2024)\cite{Liu2024-rd} \newline
    Li et al.(2024)\cite{Li2024-pj} \newline
    Lin et al.(2024)\cite{Lin2024-ly} \newline
    Beck et al.(2023)\cite{beck25sensitivity} \\
    \midrule
    \textbf{Multimodal Attacks} & FigStep \newline
    Jailbreak in Pieces \newline
    VisualJail & Typographic attacks \newline
    Image-text fusion \newline
    Cross-modal vulnerabilities \newline
    Visual prompt injection & Gong et al.(2023)\cite{Gong2023-xr} \newline
    Shayegani et al.(2023)\cite{Shayegani2023-in} \newline
    Wang et al.(2023)\cite{Wang2023-vh} \newline
    Yu et al.(2024)\cite{Yu2024-pm} \\
    \midrule
    \textbf{Defense Mechanisms} & LLM Self Defense \newline
    Moving Target Defense \newline
    APS \newline
    Erase-and-check \newline
    SafeGuard & Self-monitoring screening \newline
    Random instance selection \newline
    Lightweight classification \newline
    Safety certification \newline
    Adaptive protection \newline
    Real-time detection & Phute et al.(2023)\cite{Phute2023-ue} \newline
    Chen et al.(2023)\cite{Chen2023-eq} \newline
    Kim et al.(2023)\cite{Kim2023-mj} \newline
    Kumar et al.(2023)\cite{kumar2023certifying} \newline
    Wang et al.(2023)\cite{Wang2023-vh} \newline
    Zhou et al.(2024)\cite{Zhou2024-uh} \newline
    Shen et al.(2024)\cite{Shen2024-fe} \\
    \bottomrule
    \end{tabular}
  \end{table}

Recent technical approaches in LLM jailbreaking have demonstrated increasing sophistication and diversity. Reinforcement learning-based methods, represented by RL-JACK \cite{Chen2024-lv}, optimize attack strategies through trial-and-error interactions with target LLMs, using reward signals based on attack success rates and output toxicity levels. This black-box approach proves particularly effective against commercial LLMs where model parameters are inaccessible. Gradient-based methods like ASETF \cite{Wang2024-zh} focus on optimizing adversarial inputs through gradient computations, achieving efficient attacks by directly manipulating the embedding space while maintaining low computational requirements. Building upon these foundations, universal adversarial methods such as PRP \cite{Mangaokar2024-ii} develop attack patterns that work across multiple LLM architectures, creating universal adversarial prefixes that can be prepended to various inputs to induce harmful outputs. More sophisticated compositional attacks have emerged, breaking down harmful prompts into seemingly benign components. For instance, Prompt Packer \cite{Jiang2023-ee} embeds malicious content within complex writing tasks, DRA \cite{Liu2024-rd} employs disguised instructions, DrAttack \cite{Li2024-pj} splits harmful prompts into multiple innocent-looking sub-prompts, and ABJ \cite{Lin2024-ly} conceals attack components within analytical tasks. The frontier has further expanded to multimodal attacks, with FigStep \cite{Gong2023-xr} exploiting typographic elements in images and Jailbreak in Pieces \cite{Shayegani2023-in} combining image and text inputs to bypass traditional safety measures. In response to these evolving threats, defense mechanisms have been developed, including LLM Self Defense \cite{Phute2023-ue} for continuous self-monitoring, Moving Target Defense \cite{Chen2023-eq} for dynamic model selection, APS \cite{Kim2023-mj} for lightweight classification of potential attacks, and the Erase-and-check framework \cite{kumar2023certifying} for certified safety guarantees.

\subsection{Representative Datasets and Benchmarks}
\label{sec:datasets-sec5}

We summarize the representative datasets and
benchmarks used in the literature reviewed in this Section, including synthetic dataset construction, counter-speech generation, red-teaming, harmful-content classification, content moderation, harm prevention, and
jailbreak defense. As shown in Table~\ref{tab:section5-datasets}, the underlying resources are more heterogeneous. Some studies introduce task-specific datasets, while others evaluate on recurring benchmark suites that are reused across subareas.

\begin{table*}[t]
\caption{Representative datasets and benchmarks used by the literature reviewed in Section~5.}
\label{tab:section5-datasets}
\centering
\footnotesize
\begin{tabular}{p{2.5cm} >{\raggedright\arraybackslash}p{4.6cm} p{5.0cm}}
\toprule
\textbf{Subsection} & \textbf{Representative datasets / benchmarks} & \textbf{Primary use} \\
\midrule

Harmful Content Generation for Dataset Construction
&
OffensiveLang~\cite{das2024offensivelang}, K-OMG~\cite{392}, ParaFusion~\cite{383}, SeeGULL Multilingual~\cite{397}, KoCoSa~\cite{kim-etal-2024-kocosa}, IndicLLMSuite~\cite{394}, GPT-HateCheck~\cite{hatecheck}, CAD~\cite{283}, KoTox~\cite{264}
&
Synthetic offensive / hateful text generation, multilingual and cultural coverage, stereotype and bias analysis, sarcasm-aware toxicity, benchmark construction, and safety-alignment data creation.
\\

LLMs for Counter Speech Generation
&
DisCGen~\cite{263}, cyberbullying-intervention dialogue settings~\cite{105}, outcome-constrained counter-speech evaluation settings~\cite{hong-etal-2024-outcome}
&
Counter-speech generation, dialogue-based intervention, discourse-aware response generation, and evaluation of constructive responses to online hate.
\\

Harmful Content Generation for Red-Teaming
&
Explore--Establish--Exploit red-teaming prompt sets~\cite{27}, ARCA auditing prompt pools~\cite{123}, SAP attack-prompt datasets~\cite{Deng2023-vd}, TroubleLLM attack prompts~\cite{385}, GFlowNet-based red-teaming~\cite{520}
&
Automated red-teaming, adversarial prompt generation, harmful-behavior elicitation, and stress-testing of moderation systems.
\\

Characterizing Toxicity
&
THOS~\cite{almohaimeed2023thos}, Turkish hate speech datasets~\cite{Cam2023-nn}, cross-cultural English hate-speech annotations~\cite{lee2024exploring}, subjectivity-aware toxic comment identification data~\cite{dutta2023modeling}, implicit-hate detection benchmarks~\cite{masud2023focal}
&
Harmful-content classification, implicit hate detection, fairness analysis, prompt-based classification, fine-tuning, and synthetic augmentation.
\\

Cross-lingual Transferability for Classification
&
THOS~\cite{almohaimeed2023thos}, Turkish hate speech datasets~\cite{Cam2023-nn}, IndicLLMSuite~\cite{394}, cross-cultural English hate-speech annotations~\cite{lee2024exploring}
&
Low-resource and multilingual harmful-content classification, cross-lingual transfer, and culturally sensitive evaluation.
\\

Content Moderation
&
Multilingual harmful web-data benchmark~\cite{jansen2022perplexed}, SQuARe~\cite{lee2023square}, GPT-3 moderation explanation settings~\cite{wang2023evaluating}, multilingual learn/unlearn corpora~\cite{lu2024every}
&
Detoxification, moderation prompting, moderation explanation, multilingual moderation, and alignment-oriented moderation evaluation.
\\

Potential in Harm Prevention
&
Beavertails~\cite{ji2023beavertails}, SQuARe~\cite{lee2023square}, KoTox~\cite{16}, GenderAlign~\cite{zhang2024genderalign}
&
Safety alignment, acceptable-response generation, toxic-instruction rejection, bias mitigation, and preference-based prevention training.
\\

Defense against LLM Jailbreaking
&
Do-Not-Answer~\cite{Wang2023-vh}, multimodal jailbreak benchmarks from FigStep and Jailbreak in Pieces~\cite{Gong2023-xr, Shayegani2023-in}, jailbreak evaluation frameworks~\cite{Yu2024-pm, Zhou2024-uh}
&
Jailbreak defense evaluation, safeguard assessment, refusal robustness, multimodal safety evaluation, and mechanistic analysis of jailbreak vulnerabilities.
\\

\bottomrule
\end{tabular}

\end{table*}

Several patterns emerge from Table~\ref{tab:section5-datasets}. First, papers in Section~5.1 and
Section~5.4 are often resource-building efforts: they explicitly construct new datasets for synthetic
hate-speech generation, counter-speech, safety alignment, or prevention, such as OffensiveLang~\cite{das2024offensivelang},
KoTox~\cite{264}, Beavertails~\cite{ji2023beavertails}, SQuARe~\cite{lee2023square}, and GenderAlign~\cite{zhang2024genderalign}. Second, the
classification and moderation literature in Sections~5.2 and~5.3 tends to reuse a relatively stable
benchmark core, especially THOS~\cite{almohaimeed2023thos},
GPT-HateCheck~\cite{hatecheck}, Robbie~\cite{vogel2023explaining}, FFT~\cite{lim2023evaluating}, and multilingual harmful web-data benchmarks~\cite{jansen2022perplexed}. Third, jailbreak-defense papers in Section~5.5 increasingly converge on shared evaluation suites such as Do-Not-Answer~\cite{Wang2023-vh}, which makes cross-paper comparison more feasible.

\color{black}
 
\section{Discussion}


Our analysis reveals a rapidly evolving landscape of LLM safety challenges in harmful content generation and mitigation. The challenges progress in both nature and sophistication: from unintentional vulnerabilities rooted in model training, to deliberate exploits, to the recent emergence of self-evolving attacks. They also span multiple dimensions, from technical gaps in safety alignment to the social consequences of model bias. The pace of this evolution, especially in multi-modal integration \cite{Gong2023-xr, Shayegani2023-in} and LLM-assisted generation \cite{268, Deng2023-vd}, which suggests that current safety measures may not hold against future threats. We organize the discussion around three areas: key technical and social findings, implications for technical, industry, and policy stakeholders, and future research directions.

\subsection{Key Findings}

\textbf{Dual Nature of Challenges.}
As demonstrated in Table~\ref{tab:llm-challenges}, the challenges in controlling harmful content generation exhibit a dual nature that demands a comprehensive understanding and response. The unintentional vulnerabilities persist despite significant advances in safety alignment techniques like RLHF \cite{bai2022training} and prompt filtering \cite{kumar2023certifying}. Even the most sophisticated and well-aligned models like GPT-4, Gemini, and LLaMA demonstrate concerning patterns of systematic biases across gender, race, and cultural dimensions \cite{501, 534, 540}. Such biases are exhibited in diverse ways, from subtle linguistic preferences to more overt stereotypical associations, affecting everything from professional content generation to casual interactions \cite{91, 501, 605}. The intentional exploitation landscape presents an even more alarming picture, with attack strategies evolving at an unprecedented pace. The progression from basic text-based attacks \cite{Chen2024-lv, Wang2024-zh} to sophisticated multi-modal integration methods \cite{Gong2023-xr, Shayegani2023-in} demonstrates not only the growing complexity of security challenges but also the increasing creativity of adversarial approaches. Recent studies have shown that these advanced attack methods can achieve success rates significantly higher than traditional approaches, with some techniques increasing harmful outputs by up to sixfold \cite{66, giorgi2025humanllmbiaseshate}.

\noindent\textbf{Accelerating Technical Evolution.}
The technical sophistication of attacks has shown remarkable acceleration, following a clear evolutionary pattern that challenges existing safety paradigms. Initial text-based adversarial techniques focused on simple prompt manipulation \cite{Chen2024-lv, Wang2024-zh}, but quickly evolved to more advanced compositional methods \cite{Jiang2023-ee, Liu2024-rd} that exploit subtle weaknesses in model architectures. The emergence of multi-modal attacks represents a significant leap in attack complexity, with methods like FigStep's typographic visual attacks \cite{Gong2023-xr} and VRP's role-play based approaches \cite{ma2024-vi} demonstrating the vulnerability of cross-modal processing pipelines.

Most concerningly, the development of LLM-assisted generation methods has opened a new frontier in attack sophistication. The Tree of Attacks with Pruning (TAP) method \cite{268} has achieved unprecedented success rates of 80\% against advanced models like GPT-4 and Gemini-Pro, even in the presence of robust moderation tools. This self-improving attack capability is particularly alarming as it demonstrates how LLMs can be weaponized against their own safety measures. Furthermore, research has shown that these automated attack methods often surpass human-created prompts in their effectiveness \cite{Deng2023-vd}, suggesting a fundamental challenge to current safety paradigms.

\noindent\textbf{Interconnected Impact Dimensions.}
The challenges manifest across multiple interconnected dimensions, creating a complex web of technical, social, and security implications. In the technical dimension, fundamental limitations in model alignment persist despite significant advances in safety techniques \cite{bai2022training, kumar2023certifying}. These limitations are not merely engineering challenges but reflect deeper questions about the nature of language model training and control. Studies have shown that even models trained with explicit safety constraints can generate harmful content when presented with carefully crafted inputs \cite{31, 34, 66, 94, 108, 284, 369}.

The social dimension reveals deep-rooted issues with bias and fairness that go beyond simple content filtering. Research has demonstrated systematic biases in generated content \cite{501, 534, 540}, with particular concerns around gender representation \cite{103, 598, levy-etal-2024-gender}, racial bias \cite{91, 501, 605}, and cultural stereotypes. These biases often emerge in subtle ways, making them particularly challenging to detect and mitigate. Studies have shown that these biases can be amplified in specific contexts, such as professional writing or news generation \cite{91, 501}.

The security dimension demonstrates an evolving arms race between attack methods and defense mechanisms. Recent research has shown that attackers are increasingly leveraging the models' own capabilities for more effective exploits \cite{268, Deng2023-vd}. This includes sophisticated techniques for prompt manipulation, context poisoning, and even the generation of adversarial examples that can bypass traditional safety measures. The emergence of multi-turn dialogue exploits \cite{34, 744} and semantic conditioning techniques \cite{94, 108} suggests that attack methods are becoming increasingly sophisticated and harder to defend against.

\subsection{Implications}

The findings from our study have significant implications across multiple stakeholder groups, requiring coordinated responses at various levels:

\noindent\textbf{Technical Implications.}
Model development must undergo fundamental transformations to address the identified challenges. Current safety alignment techniques, while valuable, have shown clear limitations \cite{bai2022training, kumar2023certifying}. Future architectures need to incorporate more robust safety mechanisms that can address both unintentional biases and intentional attacks. This includes developing new approaches to model training that can handle the complexity of human values and social contexts.

The success of multi-modal attacks \cite{Gong2023-xr, Shayegani2023-in} necessitates the development of integrated defense systems that can protect against cross-modal vulnerabilities. This requires not only technical innovations in model architecture but also new approaches to data curation and training. Evaluation frameworks must expand beyond traditional metrics to assess implicit biases and potential for harmful content generation \cite{501, 534}. Recent research suggests the need for more sophisticated testing methodologies that can capture subtle forms of bias and potential for harm \cite{das2024offensivelang, hatecheck}.

Furthermore, the emergence of LLM-assisted attacks \cite{268, Deng2023-vd} calls for new paradigms in model security. Traditional approaches to content filtering and safety boundaries may no longer be sufficient when models can actively participate in their own exploitation. This suggests the need for more dynamic and adaptive security measures that can evolve alongside potential threats.

\noindent\textbf{Industry Implications.}
Companies deploying LLMs must fundamentally rethink their approach to safety and security. The demonstrated vulnerabilities in current systems \cite{28, 29} require more than incremental improvements to existing safety measures. Organizations need to implement comprehensive safety protocols that include continuous monitoring systems capable of detecting both traditional and emerging attack patterns.

The rise of sophisticated attacks like TAP \cite{268} and multi-modal exploitation methods \cite{Gong2023-xr, ma2024-vi} necessitates the development of adaptive defense mechanisms that can evolve alongside threat patterns. This includes investing in research and development of new safety technologies, as well as establishing clear protocols for incident response and mitigation.

Industry stakeholders must also address the ethical implications of their technology. Studies showing systematic biases \cite{501, 534, 540} suggest the need for more rigorous testing and validation procedures before deployment. Companies should establish clear guidelines for responsible AI deployment, incorporating lessons learned from documented exploitation cases and ensuring transparency in their safety measures.

\noindent\textbf{Policy Implications.}
The regulatory framework must evolve to address the complex challenges presented by LLMs in social media contexts. Studies showing systematic biases \cite{501, 534, 540} call for stronger oversight of model training and deployment practices. This includes developing standards for model evaluation and certification, as well as establishing clear accountability mechanisms for when systems fail to meet safety requirements.

The rapid evolution of attack methods \cite{268, Deng2023-vd} suggests the need for flexible regulatory approaches that can keep pace with technological advancement. Policy makers should focus on developing comprehensive ethical guidelines that address both current vulnerabilities and emerging threats. This includes establishing frameworks for regular safety audits and impact assessments, as well as requirements for transparency in model deployment and operation.

The demonstrated potential for societal harm through both unintentional bias \cite{501, 534} and intentional exploitation \cite{Gong2023-xr, Shayegani2023-in} underscores the importance of proactive policy measures. This includes developing guidelines for responsible AI development and deployment, establishing mechanisms for public oversight and accountability, and ensuring that regulatory frameworks can adapt to emerging challenges. Recent research on privacy preservation suggests the need for comprehensive policies that protect individual rights while enabling beneficial AI applications.

\subsection{Future work}
As the LLM landscape evolves rapidly, future efforts should address both the limitations of current safety alignment techniques and the untapped potential of LLMs in handling harmful content. We organize the future work into two major thrusts: (1) overcoming the challenges of LLMs in harmful content, and (2) amplifying the potential of LLMs in this domain.

\subsubsection{Overcoming the Challenges of LLMs in Harmful Content Generation}
\leavevmode\par 

\noindent\textbf{Leveraging LLM-Generated harmful content for Beneficial Applications:} A counterintuitive yet promising direction is the controlled use of LLM-generated harmful content for constructive purposes. Although filtering toxic content is necessary, overly restrictive moderation may hinder research and the development of effective defenses. Recent “green teaming” studies suggest that selectively bypassing filters in secure settings can yield valuable societal applications~\cite{140}.

One such application is generating synthetic harmful content datasets, which help train more robust detection models without exposing annotators to harmful real-world content. Additionally, red-teaming exercises that use controlled harmful content to probe moderation systems can uncover hidden vulnerabilities, enabling stronger alignment safeguards.

Moreover, LLMs show promise in generating counter speech—automated, context-aware responses that challenge hateful language. Future research ought to concentrate on creating adaptive counter speech models that respond dynamically to evolving harmful content patterns while adhering to ethical standards.

\noindent \textbf{Investigating the Underlying Mechanisms of Harmful Content Generation:} To develop more effective alignment techniques, it is essential to understand how and why LLMs generate harmful content. Studies indicate that current safety methods may only suppress surface-level behavior without modifying the internal representations responsible for toxicity~\cite{liu-etal-2024-intrinsic}.

Future research should investigate the effects of alignment strategies—both intrinsic (e.g., during training) and extrinsic (e.g., through RLHF)—on the model's internal token dynamics and knowledge representations. Techniques like model probing and adversarial testing can shed light on whether harmful associations are truly suppressed or merely masked. In parallel, ethical reasoning and value alignment should be embedded earlier in the model development pipeline, rather than relying solely on post-hoc interventions~\cite{691, 246}.

\noindent \textbf{Preventing and Defending Against Jailbreak Attacks:} Future work should focus on developing more robust defense frameworks that can resist advanced attack strategies. Mangaokar et al. (2024)~\cite{Mangaokar2024-ii} emphasize the need for improving guard model architectures to better handle universal adversarial prefixes. Kumar et al. (2023)~\cite{kumar2023certifying} recommend enhancing safety alignment as a potential alternative to RLHF.  With the rise of multimodal LLMs, enhancing cross-modal safety alignment is crucial. Gong et al. (2023)~\cite{Gong2023-xr} stress the importance of developing defenses for Vision-Language Models VLMs, while Shayegani et al. (2023)~\cite{Shayegani2023-in} suggest improving alignment techniques to mitigate vulnerabilities across diverse input modalities.

Advanced detection mechanisms are essential for identifying jailbreak attempts in real-time. Phute et al. (2023)~\cite{Phute2023-ue} propose integrating in-context learning and logit biasing to reinforce safety responses in LLM outputs. Kim et al. (2023)~\cite{Kim2023-mj} advocate for refining adversarial prompt datasets and improving detection robustness through better dataset curation. Improving LLMs’ understanding of user intent is a crucial defense strategy. Jiang et al. (2023)~\cite{Jiang2023-ee} recommend enhancing intent recognition and command disassembly mechanisms to counter compositional instruction attacks. Li et al. (2024)~\cite{Li2024-pj} suggest extending prompt decomposition techniques to multilingual contexts and integrating them with other adversarial frameworks for better attack prevention.

\noindent \textbf{Balancing Performance and Safety in LLMs:} A significant challenge in LLM development lies in striking an effective balance between content safety and model performance. While stricter safety mechanisms reduce the risk of generating harmful content, they often lead to diminished linguistic creativity and response diversity \cite{548}. For instance, studies on RLHF have shown that heavily aligned models frequently exhibit "mode collapse", which produces overly deterministic outputs that reduce syntactic and semantic variation. This constraint is particularly problematic in applications that demand creative or context-sensitive generation, such as dialogue systems or narrative writing.

Further compounding this issue, Bianchi et al. (2023) \cite{bianchi2023safety} observe that excessive fine-tuning for safety can result in over-cautious behavior, where models unjustifiably refuse benign prompts merely because they resemble unsafe ones. Similarly, Ma et al. (2023) \cite{ma2023adapting} find that extensive supervised fine-tuning and data engineering can cause LLMs to overreact to innocuous inputs. These findings underscore a critical usability-safety tension, where over-correction hampers legitimate functionality and undermines user trust.

To address this, future research should explore controlled text generation techniques that allow for nuanced moderation without severely restricting the expressive capabilities of LLMs. One promising direction is the use of dynamic attribute graphs (DATG), which provide a structured way to modulate the occurrence of key words while preserving generative diversity \cite{liang-etal-2024-controlled}. Such approaches can enable fine-grained safety controls that adapt to contextual nuances rather than enforcing blanket content prohibitions.

\subsubsection{Amplifying the Potentials of LLMs in Mitigating Harmful Content}
\leavevmode\par 
\noindent \textbf{Enhancing the Capability in Harmful Content Classification:} Despite notable advances in using LLMs for harmful content detection and classification, several areas demand further exploration to ensure scalable and equitable deployment. First, reducing reliance on large labeled datasets remains a priority. Future work should investigate few-shot, zero-shot, and parameter-efficient learning techniques—such as adapters, LoRA, and prompt tuning—that maintain performance under class imbalance and data scarcity~\cite{xu2024hatespeech,nasir2023llms}.

Second, harmful content often spans multiple modalities and domains. Integrating multimodal fusion and domain-adaptation techniques can help models generalize across text, image, video, and platform-specific contexts~\cite{Cam2023-nn}. Likewise, achieving multilingual fairness requires combining cross-lingual transfer with fairness-aware training to avoid performance disparities in underrepresented languages and dialects~\cite{xu2024hatespeech}.

Third, robustness against adversarial perturbations, such as manual jailbreaks and automated prompt attacks, must be improved. Hybrid defenses combining prompt engineering, adversarial training, and continual learning will be crucial to maintaining classifier integrity over time~\cite{wu2024exploring, zhou2024multimodal}.

Finally, to promote real-world adoption and trust, future systems should incorporate interpretability tools, calibrated uncertainty estimates, and transparent moderation pipelines~\cite{vogel2023explaining}. Standardized benchmarks that reflect diverse linguistic, cultural, and demographic contexts are equally important for ensuring reproducibility and ethical progress~\cite{lim2023evaluating, cao2024toxicity}.

\noindent \textbf{Enhancing the Capability in Harmful Content Prevention:} Preventing the generation of harmful content requires not only robust detection/classification systems but also proactive strategies that guide LLMs towards safer and more inclusive outputs. A growing body of research emphasizes the importance of enriched supervision to strengthen safety alignment and bias mitigation through larger, more diverse, and value-aligned pre-trained datasets \cite{xie2023empirical, ji2023beavertails, lee2023square, bang2023enabling, qureshi2024refine}.

One fundamental challenge that limits the capability of LLMs in harmful content prevention is the narrow scope of human preferences represented in existing labeled data. Ji et al. (2023) \cite{ji2023beavertails} observe that demographic homogeneity among annotators can embed a limited worldview into model training, which undermines generalization and fairness. Likewise, Lee et al. (2023) \cite{lee2023square} and Dong et al. \cite{dong2023steerlm} argue that value assumptions embedded in certain languages may constrain LLMs' normative boundaries, which limits their global applicability.

The complexity deepens when models are expected to generalize across languages and cultural domains. Bang et al. (2023) \cite{bang2023enabling} and Kaneko et al. (2024) \cite{kaneko2024evaluating} underscore the difficulty of maintaining robustness and fairness when supervision lacks cross-linguistic and cross-cultural grounding. These challenges call for inclusive training data that reflects diverse values, usage contexts, and linguistic structures. To address this, future research should pursue more comprehensive datasets that improve the granularity of safety signals and reduce blind spots in alignment. Such data enables models to make more context-aware judgments and minimize unintentional harm.

\section{Conclusions}

Through our comprehensive systematic review of LLM challenges in harmful content generation, we have illuminated a rapidly evolving landscape that demands immediate attention and action. Our analysis has uncovered a concerning progression from simple unintentional vulnerabilities to sophisticated intentional exploits, with challenges manifesting across technical, social, and ethical dimensions. While the field has made notable progress in implementing safety measures, the dynamic nature of social media environments and the emergence of advanced attack methods, particularly in multi-modal contexts, continue to present significant challenges. The gap between rapidly advancing LLM capabilities and effective safety measures remains a critical concern, highlighting the pressing need for more resilient and adaptable solutions. As we look to the future of LLM safety, success will depend on our ability to develop proactive, holistic approaches that combine technical innovation with thoughtful policy frameworks, ensuring these powerful language models can serve society while effectively mitigating their potential for harm. This will require sustained collaboration among researchers, industry stakeholders, and policymakers to create and implement multi-stakeholder solutions that can adapt to emerging threats while preserving the beneficial aspects of LLM technology.

\bibliographystyle{plain}
\bibliography{All-References}

%
%
%
%

\appendix

\section{Literature Screening and Coding Prompts}
\label{app:prompts}

To support reproducibility, this appendix documents the two prompts used
in the automated stages of the study-selection and analysis pipeline
described in Section~\ref{sec:data}. Both stages were run with
\textsf{gpt-4-turbo}, and in both stages four of the authors reviewed the
model's outputs and held final decision authority: at the abstract stage
they adjudicated the records retained for full-text retrieval, and at the
coding stage they manually verified and, where necessary, corrected the
extracted labels. The keyword groups used to build the initial query are
listed in Table~\ref{tab:keywords}.

\subsection{Abstract Relevance Screening}
\label{app:screen}

In the first stage, each of the 3{,}087 de-duplicated records was screened
at the title-and-abstract level using the prompt below. To favor recall,
the model was instructed to retain borderline records, which were then
resolved by the authors during full-text assessment.

\begin{lstlisting}[style=prompt]
You are screening academic papers for a survey on the safety of large
language models (LLMs). Given the title and abstract of a paper, decide
whether it concerns the safety or integrity of LLMs -- that is, the
generation, detection, classification, moderation, or jailbreaking of
harmful content (e.g., toxicity, hate speech, harassment, offensive or
biased language).

Answer with a single label: Include or Exclude. If the decision is
uncertain, answer Include.

Title: {title}
Abstract: {abstract}
\end{lstlisting}


\subsection{Per-Paper Aspect Coding}
\label{app:coding}

In the second stage, each of the 372 included studies was coded along the
four aspects below. The first three aspects correspond to the analytical
dimensions in Figure 2: \emph{Type of Hate Speech} to the harm-category axis and \emph{Technique or Strategy} to the
technical-approach axis. The fourth aspect, \emph{Potential and Challenges
of LLMs}, determines the ``Good'' (safety-enabling) versus ``Bad''
(harm-producing) role shown in the inner rings of Figure~\ref{fig:related studies distribution}: the labels \texttt{Promote performance}, \texttt{release the annotator}, and \texttt{efficiency of assessment} are mapped to the Guardian role,
while \texttt{Biased}, \texttt{easy to be attacked}, \texttt{unable to
identify hate speech}, and \texttt{data pollution} are mapped to the
Offender role. The LLM family (e.g., GPT, LLaMA, Claude) was recorded
separately from each paper's metadata. The harm-category labels are
multi-label: a study addressing more than one type of harm receives more
than one label, which is why the per-category counts in Figure~\ref{fig:related studies distribution} does not sum to the number of unique studies.

\begin{lstlisting}[style=prompt]
You will extract four aspects from each paper, based on its title,
abstract, and keywords. Output only JSON, with no additional text.

(1) NLP Task -- the specific NLP task of the paper.
    Options: {text generation}, {Q&A}, {text classification}.
    If none applies, choose the closest precise term from machine
    learning or NLP. Do not use broad words such as "Analysis" or
    "Others".

(2) Type of Hate Speech -- the type of hate speech the paper addresses.
    Options: {Profanity and Offensive Language},
    {Hate Speech and Discrimination}, {Harassing Behavior},
    {Content Moderation}, {Extremism and Radicalization}.
    If none applies, output "others and explain".

(3) Technique or Strategy -- the technique or strategy used.
    Options: {Training-based Method}, {Prompt Engineering}.
    If neither applies, choose a precise, consistent term that best
    describes the approach.

(4) Potential and Challenges of LLMs -- the potential or challenge the
    paper highlights for LLMs.
    Options: {Promote performance}, {release the annotator},
    {efficiency of assessment}, {Biased}, {easy to be attacked},
    {unable to identify hate speech}, {data pollution}.
    If none applies, choose the most relevant specific term(s).

Output format:
{
  "NLP Task": "...",
  "Type of Hate Speech": "...",
  "Technique or Strategy": "...",
  "Potential and Challenges of LLMs": "..."
}

Title: {title}
Abstract: {abstract}
Keywords: {keywords}
\end{lstlisting} 

\end{document}